\documentclass[smallcondensed]{svjour3}
\usepackage{listings}

\usepackage[colorinlistoftodos]{todonotes}
\usepackage{url}
\PassOptionsToPackage{hyphens}{url}
\usepackage[bookmarks=false,hypertexnames=true]{hyperref} 
\usepackage{graphicx}
\usepackage{newtxtext}
\usepackage{newtxmath}
\usepackage[round]{natbib}
\usepackage{algorithm}
\usepackage{algpseudocode}
\algloopdefx{NFor}[1]{\textbf{for all} #1 \textbf{do}}
\usepackage{tikz}
\usetikzlibrary{positioning}
\usetikzlibrary{arrows}
\usetikzlibrary{chains}

\title{High-Accuracy  Model-Based  Reinforcement Learning, \em a Survey} 
\author{Aske Plaat \and Walter Kosters \and Mike Preuss}
\date{\today}

\institute{Leiden Institute of Advanced
              Computer Science
              \\
              Leiden University, Leiden  \\
              The Netherlands \\
             \\ \email{a.plaat@liacs.leidenuniv.nl}
              }

\begin{document}
\maketitle

\begin{abstract}

Deep reinforcement learning has shown remarkable success in the past
few years. Highly complex sequential 
decision making problems from game playing and robotics
have  been solved with deep model-free methods. Unfortunately, the sample
complexity of model-free methods is often  high. To reduce the number of environment
samples, model-based reinforcement learning creates
an explicit model  of the environment dynamics.


Achieving high model accuracy is a challenge in high-dimensional
problems. 
In recent years, a diverse landscape of model-based methods has been
introduced to improve model accuracy, using methods such as
uncertainty modeling,  model-predictive 
control, latent models, and end-to-end
learning and planning.  Some of these methods succeed in achieving high
accuracy at low sample complexity, most do so either in 
 a robotics or in a games context. In this paper, we survey  
these methods; we explain in detail how they 
work and what their strengths and weaknesses are. 
We conclude
with a research agenda for future work to make the methods more robust and
more widely applicable to other applications.

\keywords{Model-based reinforcement learning
  \and
  Latent models
  \and
  Deep learning
  \and
  Machine learning
  \and Planning}
\end{abstract}

\section{Introduction}


Recent breakthroughs in game playing and robotics  have shown the
power of deep reinforcement learning, for example, by
learning to play Atari and Go from scratch or by learning to fly an acrobatic model
helicopter~\citep{mnih2015human,silver2016mastering,abbeel2007application}.  
Unfortunately, for most  applications 
the  sample efficiency is low~\citep{silver2016mastering,lecun2015deep},
and achieving faster learning is a
major topic in  current research.
%
Model-based methods can achieve faster learning 
by making an internal dynamics model of the 
environment. By then using this dynamics model for
policy updates, the number of necessary environment samples can be reduced
substantially~\citep{sutton1991dyna}.

The success of the model-based
approach hinges  on the accuracy of this
dynamics model---there is a trade-off between accuracy and sample
complexity, especially in models with many parameters that require many
samples to prevent 
overfitting~\citep{lecun2015deep,talvitie2015agnostic}. The challenge
for the methods in this survey 
is {\em   to train a high-accuracy dynamics model for
high-dimensional problems with low sample complexity}.

Reinforcement learning finds  solutions for sequential
decision problems (see Figure~\ref{fig:rltree}). 
Where model-free methods learn  the best
action in each state, 
model-based methods go a step further: the next-state transition
function in each state is  learned. Model-based methods  capture the core of complex
decision sequences, and models may also be applicable to related
environments~\citep{risi2020chess,torrado2018deep}, for transfer learning.

The  contribution of this survey is to give an in-depth
overview of recent high-accuracy model-based methods  for
high-dimensional  problems. We present  a taxonomy based on
application type, learning method, and planning method, and while
improving model accuracy is difficult,  successful methods are reported for game 
playing and visuo-motor control (although rarely both at the same time). We describe how and why the
methods work---we do note, however, that the computational cost is still high, and
that the outcomes of experiments are often sensitive to the choice of
hyperparameters.  We close with a research agenda to improve
reproducibility, to further improve accuracy, and to make methods more widely
applicable.

The field of deep model-based reinforcement learning is quite active.
Excellent works with  background
information exist for  reinforcement
learning~\citep{sutton2018introduction} and deep
learning~\citep{goodfellow2016deep}.
Previous surveys provide
an overview of the uses of classic (tabular) model-based
methods~\citep{deisenroth2013survey,kober2013reinforcement,kaelbling1996reinforcement}. The purpose of the current survey is to focus on deep learning
methods. 
Other relevant surveys into model-based reinforcement learning
are~\citep{justesen2019deep,polydoros2017survey,hui2018model,wang2019benchmarking,ccalicsir2019model,moerland2020model}. 

Section~\ref{sec:rl} provides necessary background  and the MDP 
formalism for reinforcement learning.  Section~\ref{sec:mbrl} then
surveys   the field.  Section~\ref{sec:dis} provides a discussion
reflecting on the different approaches and provides   open
problems and research directions for future work. Section~\ref{sec:con} concludes the survey.

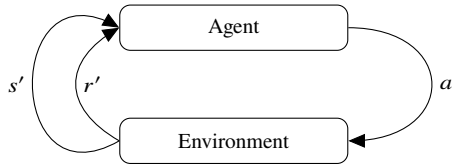
\begin{figure}[t]
  \begin{center}
    \begin{tikzpicture}[>=triangle 45,
  desc/.style={
		scale=1.0,
		rectangle,
		rounded corners,
		draw=black, 
		}]

  \node [desc,minimum width=3cm,minimum height=0.6cm] (tm) at   (0,0.5) {Environment};
  \node [desc,minimum width=3cm,minimum height=0.6cm] (pol) at   (0,2) {Agent};
  \draw (tm.west) edge[->,in=210,out=150,looseness=1.5] node[right] {$r'$} (pol.west);
  \draw (tm.west) edge[->,in=150,out=210,looseness=3] node[left] {$s'$} (pol.west);
  \draw (pol.east) edge[->,out=0,in=0,looseness=2.5] node[right] {$a$} (tm.east);

\end{tikzpicture}
  \end{center}
  \caption{Reinforcement learning: agent performs action $a$ on
    environment, which provides new state $s'$ and reward $r'$}\label{fig:agent}
\end{figure}

\section{Background}\label{sec:rl}
The goal of reinforcement learning is to learn  optimal  behavior 
for a certain environment, maximizing expected cumulative future
reward. This  goal is reached after a sequence of decisions is taken,
actions that can be different for each state;  the best sequence of
actions  solves a sequential decision making problem.
 Reinforcement learning draws
inspiration from  human and 
animal learning~\citep{hamrick2019analogues,kahneman2011thinking,anthony2017thinking,duan2016rl},
where   behavioral  adaptation  by
reward and punishment is studied.

In contrast, supervised learning methods can be used when a dataset of labeled training pairs $(x,y)$ is present. 
The reinforcement learning paradigm, however, does not assume the
presence of such a
dataset, but  derives the ground truths from  an  external
environment, see Figure~\ref{fig:agent}. The environment 
provides a new state $s'$ and its reward $r'$  for every
action $a$ that the agent tries in a certain state
$s$~\citep{sutton2018introduction}. In this way, as 
many (state-action, reward) pairs can be 
generated as needed.\footnote{A dataset is static. In reinforcement
  learning the choice of actions  may depend on the rewards that are
  returned during the learning process, giving rise to a dynamic,
  potentially unstable, learning process.}
%
%
%

\subsection{Formalizing Model-Based Reinforcement Learning}\label{sec:mdp}

Reinforcement learning problems are often modeled  as a Markov
decision process~\citep{bishop2006pattern,sutton2018introduction}, MDP. First we
introduce  state, 
action, transition and reward. Then we introduce trajectory, policy and
value. Finally, we discuss model-based and model-free solution approaches.

\begin{figure}[t]
\begin{center}
  \tikzset{
  treenode/.style = {align=center, inner sep=0pt, text centered,
    font=\sffamily},
  arn_n/.style = {treenode, circle, white, draw=black,
    fill=black, text width=2mm},
  arn_r/.style = {treenode, circle, black, draw=black, 
    text width=3mm, thick}
}

\begin{tikzpicture}[->,>=stealth',level/.style={sibling distance = 1.2cm/#1,
  level distance = 1cm}] 
\node [arn_r,label=above:{$s$}] {}
    child{ node [arn_n] {} 
            child{ node [arn_r] {} 
            }
            child{ node [arn_r] {}
            }                            
    }
    child{ node [arn_n] {} 
            child{ node [arn_r] {}
            }            
            child{ node [arn_r] {}
            }
            edge from parent node[right] {$\pi$} 
    }
    child{ node [arn_n,label=above:{$a$}] {}
            child{ node [arn_r] {} 
            }
            child{ node [arn_r,label=right:{$s'$}] {}  edge from parent node[right] {$r$} 
            }
    }
;  
\end{tikzpicture}
\caption{Backup Diagram~\citep{sutton2018introduction}. 
Maximizing  the reward for state $s$ is performed by following the {\em transition}
function to find the next state $s'$. Note that the policy $\pi(s,a)$ tells
the first half of this transition, going from $s \rightarrow a$; the
transition function $T_a(s,s^\prime)$ completes the  transition, going from $s \rightarrow
s^\prime$ (via $a$).}\label{fig:rltree}
\end{center}
\end{figure}
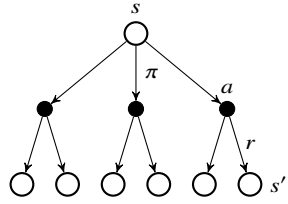

A Markov decision process is a 4-tuple $(S, A, T_a, R_a)$ where
$S$ is a finite set of states,
$A$ is a finite set of actions; $A_s \subseteq A$ is the set of actions available
from state $s$. Furthermore, $T_a$ is the transition function: $T_a(s,s')$ is
the probability that action $a$ 
in state $s$ at time $t$ will lead to state $s^\prime$ at time
$t+1$. Finally, $R_a(s,s^\prime)$ is the immediate reward received after transitioning
from state $s$ to state $s^\prime$ due to action $a$.

A policy $\pi$ is a stochastic or deterministic function mapping  states to
actions. The goal of the 
agent is to learn a policy that maximizes the value function,  the
expected cumulative discounted
sum of rewards $V(s) = \mathbb{E}_{\tau}\big[\sum_{t=0}^{T} \gamma^{t}
r_{t} \big]$ in a trajectory $\tau$, with $\gamma$ a discount
parameter in an episode with $T$ steps.   The optimal policy $\pi^\star$ contains the
solution to a sequential
decision problem: a  prescription of which action must be
taken in each state.

\begin{figure}[t]
  \begin{center}
    \begin{tikzpicture}[>=triangle 45,
  desc/.style={
		scale=1.0,
		rectangle,
		rounded corners,
		draw=black, 
		}]

  \node [desc,minimum height=0.6cm] (tm) at   (0,0.5) {Environment};
  \node [desc,minimum height=0.6cm] (pol) at   (0,2) {Policy/Value};
  \draw (tm.west) edge[->,in=180,out=180,looseness=2.5] node[left]
  {learning} (pol.west);
  \draw (pol.east) edge[->,out=0,in=0,looseness=2.5] node[right] {acting} (tm.east);

\end{tikzpicture}
    \caption{Model-Free Learning}\label{fig:free}
  \end{center}
\end{figure}
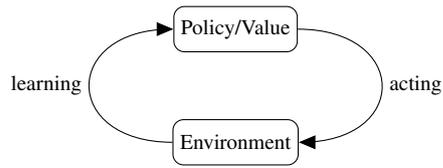

In model-free reinforcement learning only the environment knows the transition model $T_a(\cdot)$
that computes the next state $s'$ distribution; the
policy is learned directly from environment feedback $r'$  (Figure~\ref{fig:free}). In
contrast, in
model-based reinforcement learning, the agent constructs its own
version of the transition model, and the policy can be learned from
the environment feedback \emph{and} with the help of the local
transition model. 
Figure~\ref{fig:rltree} shows a visual diagram of states, actions, and transitions.

The function $V^\pi(s)$ is called the state-value function. 
Some algorithms  compute the policy $\pi$ directly, and some first compute the
function $V^\pi(s)$. For continuous or stochastic problems often  direct \emph{policy-based} methods work
best, for discrete or deterministic problems the \emph{value-based} methods are most often
used~\citep{kaelbling1996reinforcement}. A third approach combines the
best of value and policy methods:
actor-critic~\citep{sutton2018introduction,konda2000actor,mnih2016asynchronous}. In
the remainder of the paper we will see that the distinction between
continuous/discrete action spaces and policy-based/value-based
algorithms  plays an important role (see also Table~\ref{tab:overview}). 

Closely related to  the value  function is the state-action-value function
$Q^\pi(s,a)$. This $Q$-function
gives the expected sum of discounted rewards for action $a$
in state $s$, and then afterwards following policy $\pi$.  The optimal policy
can be found by recursively
choosing the argmax action with $Q(s,a)=V^\star(s)$  in 
each state. This relationship is given by $\pi^{*}= \max\limits_{\pi} V^{\pi}(s) = \max\limits_{\pi,a} Q^{\pi}(s,a)$.



There are many algorithms to find  optimal policies~\citep{hessel2017rainbow}. Algorithms that
use the agent's transition function directly to find the next state are called planning
algorithms, algorithms that use the environment to find the next state
are called learning algorithms. We now  briefly discuss
classical   model-free learning and planning approaches, 
before we continue to survey  model-based algorithms in more depth in
the next section.

In deep learning, functions such as the policy function $\pi$ are approximated
by the parameters (or weights) $\theta$  of a deep neural
network, and are written as $\pi_\theta$, to distinguish them from
classical tabular policies.

\subsection{Model-Free Learning}
When the agent does not have access to the transition
or reward model, then the policy function has to be learned by
querying the 
environment to find the reward for the action in a certain
state. Learning the policy or value 
function in this way is called model-free learning, see
Figure~\ref{fig:free}.

Recall that the policy is  a mapping of states to (best) actions. Each time when a new reward is returned by the environment
the policy can be improved: the best action for the state is
updated to reflect the new information.
Algorithm~\ref{lst:free} shows  high-level steps of model-free
reinforcement learning (later on the algorithms become more elaborate).

\begin{algorithm}[t]
    \caption{Model-Free Learning}\label{lst:free}
    \begin{algorithmic}
      \Repeat
      \State Sample env $E$ to generate data $D=(s, a, r', s')$  
      \State Use $D$ to update policy $\pi(s, a)$
      \Until $\pi$ converges
    \end{algorithmic}
\end{algorithm}

Model-free reinforcement learning is the most basic form of
reinforcement learning~\citep{kaelbling1996reinforcement,deisenroth2013survey,kober2013reinforcement}. It has been successfully applied to a range of
challenging
problems~\citep{mnih2015human,abbeel2007application}. 
In model-free reinforcement learning a policy is learned from the
ground up through interactions 
with the environment.

The  goal of classic model-free learning is to find the optimal
policy for the environment; the goal of deep model-free learning is to
find a policy function that  generalizes well to states from the
environment that have not been  seen during training. A
secondary goal is to do so with good sample efficiency: to use as
few environment samples as possible.

Model-free learning follows the current behavior policy $\pi$ in
selecting the action to try, deciding between exploration of new
actions and  exploitation of known good actions with a selection rule such as
$\epsilon$-greedy. Exploration is essentially blind, and
learning the policy and value often takes many  samples, millions in
current experiments~\citep{mnih2015human,wang2019benchmarking}.

A well-known model-free reinforcement learning algorithm is 
Q-learning~\citep{watkins1989learning}.  Algorithms such as
Q-learning were developed  in a classical tabular setting. Deep
neural networks have  been used with  success in model-free
learning, in domains in which samples can be
generated cheaply and quickly, such as in Atari video
games~\citep{mnih2015human}. Deep model-free algorithms such as Deep
Q-Network (DQN)~\citep{mnih2013playing} and
Proximal Policy Optimization, PPO~\citep{schulman2017proximal} have
become quite popular. PPO is an algorithm that computes the policy
directly, DQN finds the value function first (Section~\ref{sec:mdp}).

Model-free methods select actions in a straightforward manner, without
using a separately learned local transition model.
An advantage of the straightforward action selection is that they can
find global optima without suffering from 
selection  bias  from model imperfections. Model-based methods
may not always be able to find as good policies as model-free can.


A disadvantage is that interaction with the environment 
can  be costly. Especially when the
environment involves the real world, such as in real-world robot-interaction,
then sampling should  be minimized, for reasons of cost, and
to prevent wear of the robot arm.
In virtual environments on the other hand, model-free approaches have been quite
successful~\citep{mnih2015human}.

An overview of model-free reinforcement learning can be found
in~\citep{sutton2018introduction,ccalicsir2019model,kaelbling1996reinforcement}.

\subsection{Planning}\label{sec:planning}
\begin{figure}[t]
  \begin{center}
    \begin{tikzpicture}[>=triangle 45,
  desc/.style={
		scale=1.0,
		rectangle,
		rounded corners,
		draw=black, 
		}]

  \node [desc,minimum height=0.6cm] (tm) at   (0,0.5) {Transition Model};
  \node [desc,minimum height=0.6cm] (pol) at   (0,2) {Policy/Value};
  \draw (tm.west) edge[->,in=180,out=180,looseness=2.5] node[left] {planning} (pol.west);
  \draw (pol.east) edge[->,out=0,in=0,looseness=2.5] node[right] {acting} (tm.east);

\end{tikzpicture}
    \caption{Planning}\label{fig:plan}
  \end{center}
\end{figure}
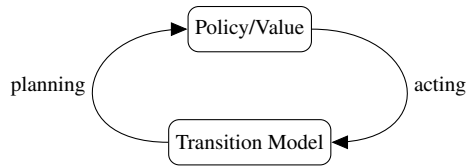

When an agent has an internal transition model, then planning
algorithms can  use it  to find the optimal policy, by
selecting actions in states, looking ahead, and backing up reward
values, see Figure~\ref{fig:rltree} and Figure~\ref{fig:plan}.
Planning algorithms require access to an explicit  model. In the deterministic case the transition model provides the next state for each of the
possible actions  in  the states, it is a function $s^\prime =
T_a(s)$. In the  stochastic case, it provides the probability
 distribution $T_a(s, s^\prime)$. The reward
model provides the immediate reward  for transitioning from state $s$
to state $s^\prime$ after taking action $a$, backing up the value from
the child state to the parent state (see the backup diagram in
Figure~\ref{fig:rltree}).
The policy function $\pi(s,a)$ concerns
the top layer of the diagram, from $s$ to $a$. The transition function
$T_a(s,s^\prime)$ covers both layers, from $s$ to $s^\prime$. 
%
The transition  function  defines a space of
states in which the planning algorithm  can search for the optimal policy $\pi^\star$ and
value $V^\star$.

\begin{algorithm}[t]
  \caption{Value Iteration}\label{lst:vi}
    \begin{algorithmic}
      \State Initialize $V(s)$ to arbitrary values
      \Repeat 
      \ForAll{$s$}
      \ForAll{$a$}
      \State $Q[s,a] = \sum_{s'} T_a(s,s')(R_a(s,s') + \gamma V(s'))$
      \EndFor
      \State $V[s] = \max_a(Q[s,a])$
      \EndFor  
      \Until V converges
      \State return V
    \end{algorithmic}
\end{algorithm}

A basic planning approach is Bellman's dynamic
programming~\citep{bellman1957dynamic}, a recursive traversal method of the
state and action space. Value iteration is a straightforward
dynamic programming algorithm. The pseudo-code for value iteration is
shown in  Algorithm~\ref{lst:vi}~\citep{alpaydin2020introduction}. It
traverses  all actions in all states,  computing the value of the
entire state space, until the value function converges.  
 


When the agent has an accurate model,  planning algorithms can
be used to find the best policy. This approach is sample efficient
since a policy is found 
without further  
interaction with the environment.
%
%
A sampling action performed in an environment is irreversible, since state changes
of the environment  can not be undone by the agent in the
reinforcement learning paradigm. In
contrast, a planning
action taken in the agent's local transition model is
reversible~\citep{moerland2020framework,moerland2018a0c}. A planning agent can 
backtrack, a sampling agent cannot. The ability to backtrack is
especially useful to try alternatives to further improve local
solutions---local solutions can be found easily
by sampling; for finding global optima efficiently the ability to backtrack
out of a local optimum improves efficiency.


\subsection{Model-Based Learning}
It is now time to look at model-based reinforcement learning, 
where the policy and
value function  are learned by both sampling and planning. Recall that the environment
samples return $(s',r')$ pairs when the agent selects action $a$ in
state $s$. This means that  we can learn the
transition model $T_a(s,s')$ and the reward model $R_a(s,s')$, for
example by supervised learning, since
all information is present. When the transition and reward model are
present in the agent, they can then be  used with planning to update
the policy and value functions as often as we like using the local
functions \emph{without any further sampling of the environment}
(although we might want to continue sampling to further improve our
models). This
alternative approach  
of finding the policy and the value is called model-based learning,
see Algorithm~\ref{lst:back} and Figure~\ref{fig:learn}.

\begin{figure}[t]
  \centering
    \begin{tikzpicture}[>=triangle 45,
  desc/.style={
		scale=1.0,
		rectangle,
		rounded corners,
		draw=black, 
		}]

  \node [desc,minimum height=0.6cm] (env) at   (4.5,0.5) {Environment};
  \node [desc,minimum height=0.6cm] (tm) at   (0.5,0.5) {Agent's Transition Model};
  \node [desc,minimum height=0.6cm] (pol) at   (2.5,2) {Policy/Value};
  \draw (env.west) edge[->,in=0,out=180,looseness=2.5] node[below]
  {learning} (tm.east);
  \draw (pol.east) edge[->,out=0,in=90,looseness=1.5] node[right] {acting} (env.north);
  \draw (tm.north) edge[->,out=90,in=180,looseness=1.5] node[left]
  {planning} (pol.west);

\end{tikzpicture}
    \caption{Model-Based Reinforcement Learning}\label{fig:learn}
  
\end{figure}
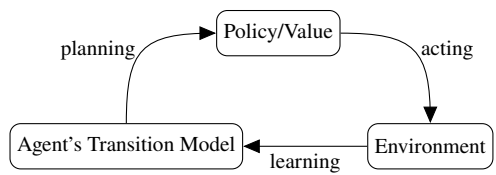

\begin{figure}[t]
  \begin{center}
    \begin{tikzpicture}[>=triangle 45,
  desc/.style={
		scale=1.0,
		rectangle,
		rounded corners,
		draw=black, 
		}]

  \node [desc,minimum height=0.6cm] (env) at   (4,0.5) {Environment};
  \node [desc,minimum height=0.6cm] (tm) at   (0,0.5) {Agent's Transition Model};
  \node [desc,minimum height=0.6cm] (pol) at   (2,2) {Policy/Value};
  \draw (env.west) edge[->,in=0,out=180,looseness=2.5] node[below]
  { learning} (tm.east);
  \draw (env.west) edge[->,in=270,out=165,looseness=1.5,thick] node[above]
  {learning} (pol.south);
  \draw (pol.east) edge[->,out=0,in=90,looseness=1.5] node[right] {acting} (env.north);
  \draw (tm.north) edge[->,out=90,in=180,looseness=1.5] node[left] {planning} (pol.west);

\end{tikzpicture}
    \caption{Dyna's Model-Based Imagination}\label{fig:imagination}
  \end{center}
\end{figure}
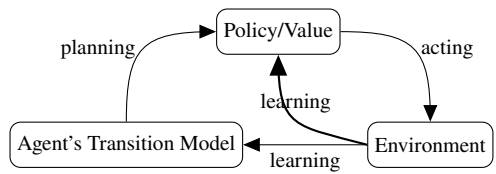

\begin{algorithm}[t]
  \begin{algorithmic}
    \Repeat
    \State Sample env $E$ to generate data $D=(s, a, r', s')$ 
    \State Use $D$ to learn $T_a(s,s')$ and  $R_a(s,s')$ 
    \State Use $T, R$ to update policy $\pi(s, a)$ by planning 
    \Until $\pi$ converges
  \end{algorithmic}
  \caption{Model-Based Reinforcement Learning}\label{lst:back}
\end{algorithm}

Why would we want to go this convoluted learning-and-planning route,
which may even introduce model-bias, if
the environment samples can teach us the optimal policy and value directly? The
reason is that the convoluted route may be more sample efficient.
In model-free learning
a sample is used once to optimize the policy, and then thrown away, in
model-based learning 
the sample is used to learn a transition model, which can then be used many
times in planning to optimize the policy. The sample is
used more efficiently by the agent.
 
A well-known classic model-based approach is \emph{imagination}, which was
introduced by~\citet{sutton1990integrated,sutton1991dyna} 
in the Dyna system, long before deep learning was used widely. Dyna
uses the samples to update the policy 
function directly (model-free learning) and also uses the 
samples to learn a transition model,  to 
augment the model-free environment-samples with the model-based
imagined ``samples.'' 
Imagination is a hybrid algorithm that
uses both model-based planning and model-free learning to improve the
behavior policy.
Figure~\ref{fig:imagination} illustrates
the working of the Dyna approach.
Algorithm~\ref{lst:dyna} shows the
steps of the algorithm (compared to  Algorithm~\ref{lst:back}, the
line in italics is new, from Algorithm~\ref{lst:free}). Note how the
policy is updated twice in each iteration, by environment sampling, and by
transition planning. More details are shown in Algorithm~\ref{alg:dyna-q}~\citep{sutton1990integrated}.

\begin{algorithm}[t]
  \begin{algorithmic}
    \Repeat
    \State Sample env $E$ to generate data $D=(s, a, r', s')$ 
    \State {\em Use $D$ to update policy $\pi(s, a)$}
    \State Use $D$ to learn $T_a(s,s')$ and  $R_a(s,s')$ 
    \State Use $T, R$ to update policy $\pi(s, a)$ by planning 
    \Until $\pi$ converges
    \end{algorithmic}
    \caption{Dyna's Model-Based Imagination}\label{lst:dyna}
\end{algorithm}

%


\begin{algorithm}[t]
  
        \begin{algorithmic}
            \State Initialize $Q(s,a) \rightarrow \mathbb{R}$ randomly
            \State Initialize $M(s,a) \rightarrow \mathbb{R}\times S$ randomly \Comment{Model}
            \Repeat
                \State Select $s \in S$ randomly
                \State $a \gets \pi(s)$ \Comment{$\pi(s)$ can be
                  $\epsilon$-greedy$(s)$ based on $Q$}
                \State $(s',r) \gets E(s, a)$ \Comment{Learn new
                  state and reward from environment}
                \State $Q(s, a) \gets  Q(s, a) + \alpha \cdot [r +
                \gamma \cdot \max_{a'} Q(s', a') - Q(s,a)]$
                \State $M(s, a) \gets (s', r)$
                \For{$n = 1, \dots, N$}
                    \State Select $\hat{s}$ and $\hat{a}$ randomly
                    \State $(s', r) \gets M(\hat{s}, \hat{a})$
                    \Comment{Plan imagined state and reward from  model}
                    \State $Q(\hat{s}, \hat{a}) \gets  Q(\hat{s}, \hat{a}) + \alpha \cdot [r + \gamma \cdot \max_{a'} Q(s', a')-Q(\hat{s},\hat{a})]$
                \EndFor
            \Until $Q$ converges
        \end{algorithmic}

  \caption{Dyna-Q: Classic learning and planning with a Q-function-based dynamics model~\citep{sutton1990integrated}} 
  \label{alg:dyna-q}
\end{algorithm}

After these introductory words, we are now ready to take a deeper look
into recent concrete deep model-based reinforcement learning methods.

\section{Survey of Model-Based Deep Reinforcement Learning}\label{sec:mbrl}

The success of model-based reinforcement learning in high-dimensional
problems depends  on the accuracy of the dynamics model.
The model is typically used by planning
algorithms for multiple sequential predictions, and errors in
predictions accumulate quickly with each step. 
%
Model-based reinforcement learning is an active field, and many
papers have been published that document progress towards improving
model-accuracy. The methods that are proposed in the papers are quite
diverse.

We will now present
our taxonomy. The taxonomy distinghuishes three aspects: 
(1) {\em application} type,  (2) {\em learning} method, and (3) {\em planning}
method. Table~\ref{tab:tax} summarizes the taxonomy,
which is the basis of the remainder of this survey. We will now
describe the methods, explaining how they fit together by going through the  learning,
planning, and application that they use.

\begin{table}[t]
  \begin{center}\footnotesize
    \begin{tabular}{lll}
      {\bf Application}&{\bf Learning}&{\bf Planning}\\
      \hline\hline
      {Discrete}   &  CNN/LSTM              & End-to-end learning and planning\\
                   &  Latent models         & Trajectory rollouts\\ \hline
      {Continuous} & Latent Models          & Trajectory rollouts\\
                   & Uncertainty  modeling  & Model-predictive control\\
                   & Ensemble models        & \\
      \hline
      
    \end{tabular}
    \caption{Taxonomy:
      {\em Application, Learning, Planning}}\label{tab:tax}
  \end{center}
\end{table}

First, we will describe the way in which the 
model is \emph{learned}, and how the accuracy of the model is
improved. Among the approaches are uncertainty modeling such as Gaussian
processes and ensembles, and convolutional neural networks or latent
models (sometimes  integrated in end-to-end learning and planning).

Second, we will describe the way in which the model is subsequently
used by the \emph{planner} to improve the behavior policy
(Figure~\ref{fig:imagination}). These 
methods aim to reduce the impact of planning with inaccurate
models. Among the methods are planning with (short) trajectories,
model-predictive control, and end-to-end learning and planning.

Finally, we will describe \emph{applications} on which model-based methods have
been successful. 
The effectiveness of model-based methods depends on whether they fit the application
domain in which they are used, and on further aspects of the
application.
There are two main types of applications, those with continuous action
spaces, and those with discrete action spaces. 
For continuous action spaces, simulated physics robotics in MuJoCo is a
favorite test bed~\citep{todorov2012mujoco}. 
For discrete action spaces many
researchers use mazes or blocks puzzles. For large, high dimensional, problems the Arcade Learning
Environment is used, where the
input consists of the screen 
pixels, and the output actions are the joystick movements~\citep{bellemare2013arcade}. 
We will use this taxonomy to categorize and understand the recent
literature on high-accuracy model-based reinforcement learning. We
list some of the papers in Table~\ref{tab:overview},
\begin{table*}[t]
  \begin{center}
    \begin{tabular}{llll}
      {\bf Name}&{\bf Learning}&{\bf Planning}&{\bf Application}\\
      \hline\hline
      PILCO~\citep{deisenroth2011pilco}&Uncertainty& Trajectory &Pendulum\\
      iLQG~\citep{tassa2012synthesis}& Uncertainty &MPC& Small\\
      GPS~\citep{levine2014learning}& Uncertainty& Trajectory&Small\\
      SVG~\citep{heess2015learning}&Uncertainty& Trajectory& Small\\
       Local Model~\citep{gu2016continuous}& Uncertainty& Trajectory& MuJoCo\\
      Visual Foresight \citep{finn2017deep}& Video Prediction&MPC&Manipulation\\
      PETS \citep{chua2018deep}& Ensemble&MPC& MuJoCo\\
      MVE  \citep{feinberg2018model}  & Ensemble& Trajectory& MuJoCo\\
      Meta Policy \citep{clavera2018model}&Ensemble& Trajectory & MuJoCo\\
      Policy Optim  \citep{janner2019trust}&Ensemble& Trajectory &MuJoCo\\
              PlaNet \citep{hafner2018learning}&Latent& MPC&MuJoCo\\
              Dreamer \citep{hafner2019dream}&Latent& Trajectory&MuJoCo\\
              Plan2Explore
      \citep{sekar2020planning}&Latent&Trajectory&MuJoCo\\
      \hline
              Video-prediction \citep{oh2015action}&Latent&Trajectory&Atari\\
              VPN \citep{oh2017value}&Latent&Trajectory &Atari\\
              SimPLe \citep{kaiser2019model}&Latent&Trajectory&Atari\\
              Dreamer-v2 \citep{hafner2020mastering}&Latent& Trajectory&Atari\\
      MuZero \citep{schrittwieser2020mastering}&Latent&e2e/MCTS&Atari/Go\\
VIN \citep{tamar2016value}&CNN&e2e&Mazes\\
      VProp \citep{nardelli2018value}& CNN&e2e&Mazes\\
    Planning \citep{guez2019investigation}&CNN/LSTM&e2e&Mazes\\
     TreeQN \citep{farquhar2018treeqn}&  Latent& e2e&Mazes\\
      I2A \citep{racaniere2017imagination}&Latent&e2e&Mazes\\
              Predictron \citep{silver2017predictron}& Latent&e2e&Mazes\\
              World Model \citep{ha2018world}&Latent & e2e&Car Racing\\
      \hline\hline\\
      
    \end{tabular}
    \caption{Overview of High-Accuracy Model-Based Reinforcement
      Learning Methods; Top: Continuous/Policy-based, Bottom: Discrete/Value-based}\label{tab:overview}
  \end{center}
\end{table*}
which provides an overview of many of the
methods that we  discuss in this survey. 
We will explain the main issues and challenges in the field step by
step, using the taxonomy as guideline, illustrating solutions to
these issues and challenges with approaches from the papers from the
table.

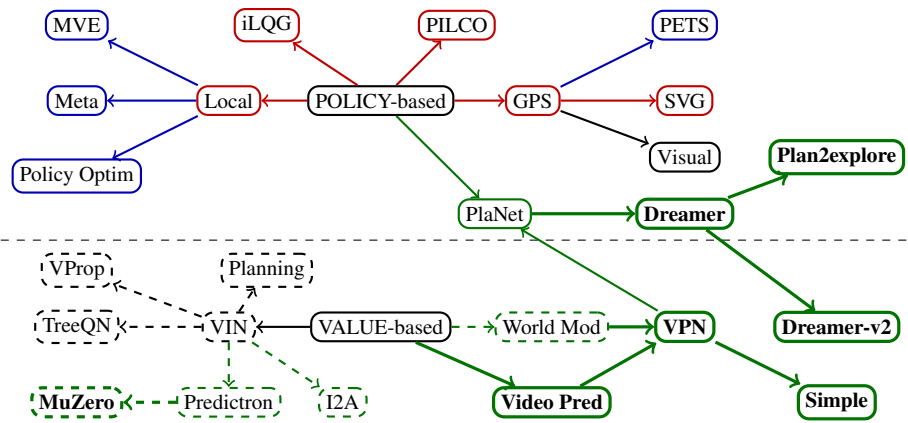
\begin{figure}
   \begin{center}
   \begin{tikzpicture}[->,scale=1,font=\footnotesize,
  desc/.style={ 
		scale=1.0,
		rectangle,
		rounded corners,
                draw=black
		},
  descg/.style={
		scale=1.0,
		rectangle,
		rounded corners,
                draw=black!55!green
		},
  descy/.style={
		scale=1.0,
		rectangle,
		rounded corners,
                draw=black!30!blue
		},
  descr/.style={
		scale=1.0,
		rectangle,
		rounded corners,
                draw=black!25!red
              }]
              
  \node[desc,thick] (value)               at (4cm,1cm)    {VALUE-based};
  \node[desc,dashed,thick] (vin)          at (2cm,1cm)    {VIN};
  \node[descg,dashed,thick] (i2a)         at (3.5cm,0cm)  {I2A};
  \node[desc,dashed,thick] (planning)     at (2.5cm,1.75cm) {Planning};
  \node[descg,dashed,thick] (predictron)  at (2cm,0cm)    {Predictron};
  \node[descg,dashed,very thick] (muzero) at (0cm,0cm)    {\bf MuZero};
  \node[desc,dashed,thick] (vprop)        at (0cm,1.75cm) {VProp};
  \node[desc,dashed,thick] (treeqn)       at (0cm,1cm)    {TreeQN};

  \node[descg,thick,dashed] (world)   at (6.25cm,1cm)     {World Mod};
  \node[descg,very thick] (video)     at (6.25cm,0cm)     {\bf Video Pred};
  \node[descg,very thick] (vpn)       at (8cm,1cm)        {\bf VPN};
  \node[descg,very thick] (simple)    at (10cm,0cm)       {\bf Simple};
  \node[descg,thick]      (planet)    at (5.5cm,2.5cm)    {PlaNet};
  \node[descg,very thick] (dreamer)   at (8cm,2.5cm)      {\bf Dreamer};
  \node[descg,very thick] (dreamerv2) at (10cm,1cm)       {\bf Dreamer-v2};
  \node[descg,very thick] (plan2)     at (10cm,3.25cm)    {\bf Plan2explore};

  \node[desc,thick] (policy)          at (4cm,4cm)    {POLICY-based};  
  \node[descr,thick] (local)          at (2cm,4cm)    {Local};  
  \node[descy,thick] (mve)            at (0cm,5cm)    {MVE};  
  \node[descy,thick] (meta)           at (0cm,4cm)    {Meta};  
  \node[descy,thick] (optim)          at (0cm,3cm)    {Policy Optim};  
  \node[descr,thick] (pilco)          at (5cm,5cm)    {PILCO};
  \node[descr,thick] (ilqg)           at (2.5cm,5cm)   {iLQG};

  \node[descr,thick] (gps)            at (6cm,4cm)     {GPS};  
  \node[descy,thick] (pets)           at (8cm,5cm)     {PETS};  
  \node[descr,thick] (svg)            at (8cm,4cm)     {SVG};  
  \node[desc,thick]  (visual)         at (8cm,3.25cm)  {Visual};  

\draw[-,black,dashed] (-1cm,2.15cm) to (11cm,2.15cm);
  
   \draw[black!55!green,->, thick,black]  (value) to  (vin);
   \draw[black!55!green,->, thick,dashed]  (vin) to  (i2a);
   \draw[black!55!green,->, thick,dashed,black]  (vin) to  (planning);
   \draw[black!55!green,->, thick,dashed]  (vin) to  (predictron);
   \draw[black!55!green,->, thick,dashed,black]  (vin) to  (vprop);
   \draw[black!55!green,->, thick,dashed,black]  (vin) to  (treeqn);
   \draw[black!55!green,->,very thick,dashed]  (predictron) to  (muzero);

   \draw[black!55!green,->, thick,dashed]  (value) to  (world);
   \draw[black!55!green,->,very thick]  (value) to  (video);
   \draw[black!55!green,->,very thick]  (world) to  (vpn);
   \draw[black!55!green,->,very thick]  (video) to  (vpn);
   \draw[black!55!green,->,very thick]  (vpn) to  (simple);
   \draw[black!55!green,->, thick]  (vpn) to  (planet);
   \draw[black!55!green,->,very thick]  (planet) to  (dreamer);
   \draw[black!55!green,->,very thick]  (dreamer) to  (dreamerv2);
   \draw[black!55!green,->,very thick]  (dreamer) to  (plan2);

   \draw[black!25!red,->, thick]  (policy) to  (local);
   \draw[black!30!blue,->, thick]  (local) to  (meta);
   \draw[black!30!blue,->, thick]  (local) to  (mve);
   \draw[black!30!blue,->, thick]  (local) to  (optim);
   \draw[black!25!red,->, thick]  (policy) to  (pilco);
   \draw[black!25!red,->, thick]  (policy) to  (ilqg);
   \draw[black!55!green,->, thick]  (policy) to  (planet);

   \draw[black!25!red,->, thick]  (policy) to  (gps);
   \draw[black!30!blue,->, thick]  (gps) to  (pets);
   \draw[black!25!red,->, thick]  (gps) to  (svg);
   \draw[black,->, thick]  (gps) to  (visual);

\end{tikzpicture}
  \caption{Influence of Model-Based Deep Reinforcement Learning
    Approaches; Top:
    Continuous/Policy-based (MuJoCo), Bottom: Discrete/Value-based
    (Mazes, Atari);  Red: Uncertainty, Blue: Ensemble, Green: Latent Models,
    Dashed: end-to-end, Bold: Large Problems }\label{fig:influence}
  \end{center}
  \end{figure}

Figure~\ref{fig:influence} illustrates how the approaches of the
papers  influence each other. Note that, as is often the case in reinforcement
learning, the influence has two origins: policy-based methods for
continuous action spaces (robotics, upper part), and value-based methods for
discrete action spaces (games, lower part). The colors in the figure refer to
approaches that are also listed in Table~\ref{tab:overview}.

Let us now start with the taxonomy. We begin with \emph{learning},
next is \emph{planning}, and finally \emph{applications}.

\subsection{Learning}
The transition model is what gives model-based reinforcement learning
its name. The accuracy of the model is of great importance, planning
with inaccurate
models will not improve the policy much,  planning with a biased
model will even harm the policy, and performance of model-based
methods will be worse than the model-free baseline~\citep{gu2016continuous}. Getting
high-accuracy models with few samples is challenging when the model has many parameters,
since, in order to prevent overfitting, we would need many environment
observations (high sample complexity). 

In this section we will describe techniques that have been developed
to improve model accuracy. We will discuss:
\begin{enumerate}
\item uncertainty modeling,
\item ensemble methods, 
\item latent models.
\end{enumerate}
For environments with continuous action spaces and non-determinism,
such as robotics, uncertainty modeling and ensembles have shown
progress. Latent models were developed in both  continuous
and  discrete action spaces.

\subsubsection{Uncertainty Modeling}
One of the shortcomings of conventional reinforcement learning methods
is that they only focus on expected value, ignoring the variance of
values. This is  problematic when few samples are taken for each
trajectory $\tau$. Uncertainty modeling methods have been developed
to counter this problem. Gausian processes can learn simple processes
with good sample efficiency, although for high-dimensional problems
they need many samples. They have been used for probabilistic inference to learn
control~\citep{deisenroth2011pilco} in the PILCO system. This system
was effective on  Cartpole and Mountain car
(Figure~\ref{fig:cartpole4}), but does not scale to larger problems.


A related method uses
nonlinear least-squares optimization~\citep{tassa2012synthesis}. Here
the model learner uses quadratic 
approximation on the  reward function, which is then used with linear approximation of the
transition function. With further enhancements this method was able to teach a humanoid robot
 how to stand up (see Figure~\ref{fig:dcs}).

We can also sample from a trajectory distribution optimized
for cost, and to
use that to train the policy, with a policy-based
method~\citep{levine2013guided}. Then we can  optimize
policies with the aid of  locally-linear models and a 
stochastic trajectory optimizer.
This approach, called Guided policy search (GPS), has been shown to train
complex policies with thousands of 
parameters learning tasks in  MuJoCo such as  swimming, hopping
and walking.
Alternatively, we can compute value gradients along the real environment trajectories,
instead of planned ones, and re-parameterize the trajectory through
sampling, to mitigate learned
model inaccuracy~\citep{heess2015learning}. This was done by
Stochastic value gradients (SVG) with global
neural network  value function approximators.

Learning arm and hand manipulation directly
from video camera input is a challenging problem in robotics. 
The camera image provides a high dimensional 
 input and increases problem size and complexity of the subsequent
 manipulation task substantially.
 Both \citet{finn2017deep,ebert2018visual} introduce 
a method called  Visual
foresight.
This system uses a 
training procedure where data is
sampled according to a probability distribution. Concurrently, a 
video prediction model is trained. This model
generates a sequence of future frames based on an
image and  a sequence of actions, as in GPS.
At test time, the least-cost sequence of actions is selected in  a
model-predictive control planning framework (see Section~\ref{sec:mpc}).
%
This approach is able to  perform
multi-object manipulation, pushing, picking and placing, and
cloth-folding tasks (which adds the difficulty of material that
changes shape as it is being manipulated).

\subsubsection{Ensembles}
Ensemble methods, such as a random forest of decision trees \citep{ho1995random}, are widely used in machine
learning~\citep{bishop2006pattern}, and they are also used in
controlling uncertainty in  high dimensional
modeling. Ensemble methods mitigate variance and improve
performance by running algorithms multiple times.  They are used with success in  model-based deep
reinforcement learning as well. \citet{chua2018deep}  combine
uncertainty-aware modeling   with sampling-based uncertainty
propagation, creating a method called Probabilistic ensembles with
trajectory sampling, PETS. (This approach is  described in the next
section, see Algorithm~\ref{alg:pets}). An ensemble of probabilistic
neural network models is used by~\citet{nagabandi2018neural}. Ensembles perform
well; performance on pusher, reacher, and half-cheetah (see
Figure~\ref{fig:dcs}) is reported to
approach asymptotic model-free baselines such as
PPO~\citep{schulman2017proximal}.
Ensembles of probabilistic networks~\citep{chua2018deep} are also used
with short rollouts, where the model horizon is shorter than the
task horizon~\citep{janner2019trust}. Results have been reported
for hopper, walker, and half-cheetah, again matching the performamce of
model-free approaches. 

The ensemble approach is related to meta learning, where we try to
speed up learning  a new task by learning from previous, related,  tasks~\citep{brazdil2008metalearning,hospedales2020meta,huisman2021survey}.  MAML is a popular
meta learning approach~\citep{finn2017model}, that attempts to learn
a network initialization $x$ such that for any task  $M_k$  the policy
attains  maximum performance  after one policy
gradient step. The MAML approach can be used to improve model accuracy by
learning an ensemble of dynamics models and by then meta-optimizing the policy
for adaptation in each of the learned
models~\citep{clavera2018model}. Results  indicate that such
meta-learning of a policy over an
ensemble of learned models indeed approaches  the level of performance of
model-free methods with substantially better sample complexity.

\subsubsection{Latent Models}
The next group of methods that we describe are the latent 
models. 
Central to all our approaches is the need for improvement of model
accuracy in complex, high-dimensional, problems.
The main challenge to achieve high accuracy
is to overcome the size of the high-dimensional state space. The idea
behind latent models is that in most high-dimensional environments
there are elements that are less important, such as background trees
that never move,  that have little or no relation with the reward
of the agent's actions. The goal of latent models is to
abstract away these  unimportant elements of the input space, 
reducing the effective dimensionality of the space. They do so by
learning the relation between the elements of the input and the
reward. When we focus
our learning mechanism  on the changes in observations that are
correlated with changes in these values, then we can improve the
efficiency of  learning high-dimensional problems greatly.
Latent models thus learn a smaller representation, smaller than the
observation space.
Planning  takes place in this smaller representation space.

\begin{figure}[t]
 \begin{center}
 \includegraphics[width=7cm]{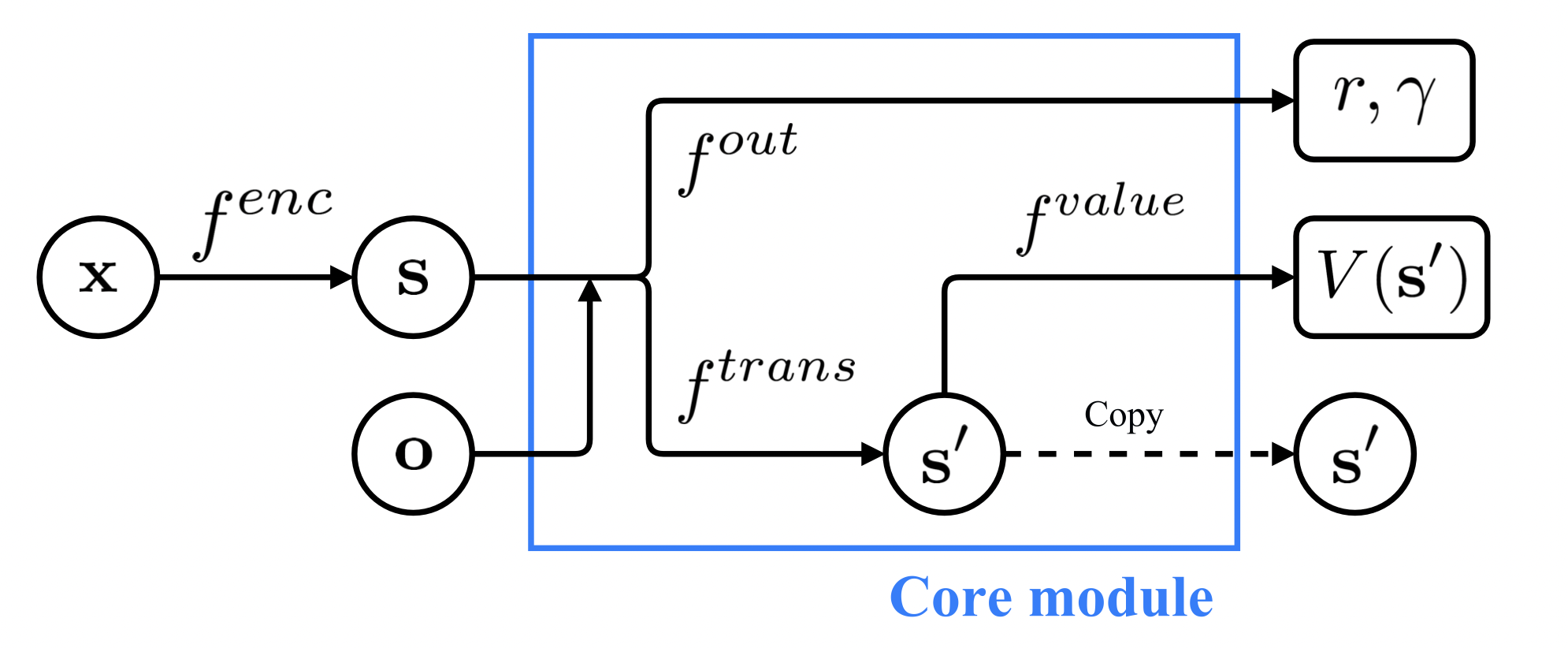}
 \caption{Architecture of latent model~\citep{oh2017value}}\label{fig:oh}
 \end{center}
 \end{figure}

The  value prediction network (VPN) was introduced by~\citet{oh2015action,oh2017value} 
to achieve this goal. They ask the question in their paper: ``What if
we could predict future rewards and values directly without predicting
future observations?'' and describe a nework architecture and learning
method for such focused value prediction models. The core idea is not to
learn directly in actual observation space, but 
first to
transform the actual state respresentation to a smaller latent representation
model, also known as abstract model. The other functions, such as
value, reward, and next-state, then work with the smaller latent representations,
instead of the actual high-dimensional states. By training all functions based on the
values~\citep{grimm2020value}, planning and learning occur in a space
where states are encouraged only to contain the elements that influence
value changes. 
In VPN the latent model consists of four networks: an encoding
function, a reward function, a value function, and a transition
function. All functions are parameterized with their own set of
parameters (Figure~\ref{fig:oh}).
Latent space
is lower-dimensional, and training and planning become more efficient.
The figure shows a single step rollout, planning one step ahead, 
as in Dyna-Q (Algorithm~\ref{lst:dyna}).



The training of the networks can in principle be performed with any
value-based reinforcement learning algorithm. \citet{oh2017value} report results
with $n$-step Q-learning and temporal difference search~\citep{silver2012temporal}.

VPN~\citep{oh2017value} showed impressive results on Atari games such as Pacman
and Seaquest, outperforming model-free DQN~\citep{mnih2015human}, and outperforming
observation-based planning in stochastic domains.
Subsequently, many other works have been published that further
improved
results~\citep{kaiser2019model,hafner2018learning,hafner2019dream,hafner2020mastering,sekar2020planning,silver2017predictron,ha2018world}.
Many of these latent-model approaches are complicated designs, with
multiple neural networks, and different learning and planning algorithms.


The latent-model approach is related to world models, a term used
by~\citet{ha2018recurrent,ha2018world}. 
World models are inspired
by the manner in which humans are thought to contruct a mental model of
the world in which we live. World models are often
generative recurrent neural networks that are trained unsupervised  using a variational
autoencoder~\citep{kingma2013auto,kingma2019introduction,goodfellow2014generative} and a 
recurrent network. They  learn a compressed spatial and temporal
representation of the environment. In world models  multiple neural
networks are used, for  a vision model, a
memory model, and a controller~\citep{ha2018world}. By using features extracted from
the world model as inputs to 
the agent, a compact and simple policy can be trained to solve a
task, and planning occurs in the compressed or simplified world. World
models have been applied by~\citet{ha2018recurrent,ha2018world} on a
car racing game~\citep{kempka2016vizdoom}.
The term
\emph{world model} actually goes back to 1990, where it was used
by~\citet{schmidhuber1990making}. Latent models are also related to dimensionality
reduction~\citep{van2009dimensionality}.
The architecture of latent models, or world models, is elaborate. The
dynamics model typically includes an observation model, a
representation model, a transition model, and a value or reward
model~\citep{karl2016deep,buesing2018learning,doerr2018probabilistic}. The task of the observation model is to reduce the
high-dimensional world into a lower-dimensional world, to allow more
efficient planning. Often a variational autoencoder or LSTM is used. 

The Arcade learning environment is one of the main benchmarks in
reinforcement learning. The 
high-dimensionality of Atari video input has long been problematic for
model-based reinforcement learning. Latent models were instrumental in
reducing the dimensionality of Atari, producing the first successes
for model-based approaches on this major benchmark.

Related to the VPN approach~\citep{oh2015action,oh2017value} is
other work that uses latent models on Atari, such
as~\citet{kaiser2019model}, that is aimed at video prediction,
outperforming model-free baselines~\citep{hessel2017rainbow}, reaching
comparable accuracy with up to an order of magnitude better sample
efficiency. The approach by~\citet{kaiser2019model} uses a variational
autoencoder (VAE) to process input frames, conditioned on the actions of the
agent, to learn the world model, using PPO~\citep{schulman2017proximal}. The policy $\pi$ is then improved by planning inside the
reduced world model, with short rollouts. This behavior policy $\pi$ then determines the
actions $a$ to be used for learning from the environment.

Latent models are also used on continuous MuJoCo problems. Here the
work by~\citet{hafner2018learning,hafner2019dream} on the PlaNet and
Dreamer systems is
noteworthy including the
application of their work back to Atari~\citep{hafner2020mastering},
which achieved human-level performance.
PlaNet uses a Recurrent state space model (RSSM) that consists of a transition model, an observation model, a variational
encoder and a
reward model~\citep{karl2016deep,buesing2018learning,doerr2018probabilistic}. 
Based on these models a Model-predictive control agent
is used to adapt its plan, replanning each
step~\citep{richards2005robust}. The RSSM is 
used by a Cross entropy method search (\citet{botev2013cross}, CEM)  for the best action
sequence. 
In contrast to model-free approaches, no explicit policy or value
function network is used; the policy is implemented as MPC planning
with the best sequence of future actions. PlaNet is tested on
continuous tasks and reaches performance that is close to
strong model-free algorithms. A further system, called
Dreamer~\citep{hafner2019dream}, builds on PlaNet. Using an actor
critic approach~\citep{mnih2016asynchronous} 
and backpropagating value gradients through predicted sequences of compact
model states, the improved system  solves a diverse collection of
continuous problems from the Deepmind control
suite~\citep{tassa2018deepmind}, see Figure~\ref{fig:dcs}. Dreamer is also applied to discrete 
problems from the Arcade learning environment, and to few-shot
learning~\citep{sekar2020planning}. A further improvement 
achieved human-level performance on 55 Atari games, a first for a
model-based approach~\citep{hafner2020mastering}, showing that the
latent model approach is well-suited for high-dimensional problems.



\subsection{Planning}
After the transition model has been learned, it will be used with a
planning algorithm so that the behavior policy can be improved. 
Since the  transition model will contain some inaccuracy, the
challenge is to find a planning algorithm that performs well  despite
the inaccuracies.
We describe three groups of methods that have been developed for planning
algorithms to cope with inaccurate models. These are:
\begin{enumerate}
\item trajectory rollouts,
\item model-predictive control,
\item end-to-end learning and planning.
\end{enumerate}
Trajectory rollouts and model-predictive control have been
shown to work for both continuous and discrete action spaces; end-to-end learning and planning
has been developed in the context of discrete action spaces (mazes and
games).

Of the three planning methods, we will start with these trajectory rollouts.


\subsubsection{Trajectory Rollouts}

%
%
As we saw in Section~\ref{sec:mdp}, methods for continuous action spaces
typically sample full trajectory rollouts to get  stable actions.
At each planning step,  the transition
model $T_a(s) \rightarrow s'$ computes the new state, using the
reward to update the policy.
Due
to the inaccuracies of the internal model,  planning algorithms  that
perform many steps will quickly accumulate model
errors~\citep{gu2016continuous}.
Full  rollouts  of long and inaccurate trajectories are therefore problematic.
We can  reduce the  impact of accumulated model errors by not  planning too far ahead. For example,~\citet{gu2016continuous} perform experiments with locally linear
models that roll out planning trajectories of length 5 to 10. This reportedly works well for
MuJoCo tasks gripper and reacher. 

In their work on model-based value
expansion (MVE),~\citet{feinberg2018model} also allow imagination to fixed
depth, value estimates 
are split into a near-future model-based
component and a distant future model-free component. They experiment
with model horizons of 1, 2, and 10. They find that 10 generally performs best
on typical MuJoCo tasks such as swimmer, walker, and cheetah. The
sample complexity in their experiments is better than model-free methods such as
DDPG~\citep{silver2014deterministic}. 
Similarly good results are reported
by~\citet{janner2019trust,kalweit2017uncertainty}, both approaches use a model horizon that is
much shorter than the task horizon.




\subsubsection{Model-Predictive Control}\label{sec:mpc}
Taking the idea  of shorter trajectories for planning than for
 learning further, we arrive at Model-predictive
control
(MPC)~\citep{kwon1983stabilizing,garcia1989model}. Model-predictive
control is a well-known approach  in 
process engineering, to control complex processes with
frequent re-planning of a limited time horizon. Model-predictive
control  uses the fact that while many real-world processes are
not linear, they are approximately linear over a small operating
range. Applications are found in the automotive industry and in
aerospace, for example for terrain-following and obstacle-avoidance
algorithms~\citep{kamyar2014aircraft}.  In optimal control,
four MPC approaches are identified: linear model MPC, nonlinear prediction model, explicit
control law MPC, and
robust MPC to deal with disturbances~\citep{garcia1989model}. In this survey, we focus on how the principle of
continuous replannning with a rolling planning horizon performs  in nonlinear
model-based reinforcement learning.

It is instructive to compare the MPC and linear quadratic regulators
(LQR) approach, since both methods  come from the field of optimal control.
MPC computes the target function with a small time window that rolls
forward as new information comes in; it is dynamic. LQR computes the
target function in a single episode, using all available information; it
is static. We observe that in model-based reinforcement learning MPC is used
in the planning part with the behavior policy $\pi$ being the target and the
transition function $T_a(\cdot)$ the input; for LQR the transition
function $T_a(\cdot)$ is the 
target, and the environment samples $(s_t, r_t)$ are the input. Thus,
one could conceivably use both MPC and LQR, the first  as planning
and the second as
learning algorithm, in a model-based approach.

An iterative form of LQG has indeed been used together with MPC on a smaller MuJoCo
problem~\citep{tassa2012synthesis}, achieving good results. MPC used
step-by-step real-time local 
optimization; \citet{tassa2012synthesis} used many  further improvements to the trajectory
optimization, physics engine, and cost function to achieve good performance.

MPC has also been
used in other model learning
approaches. Both~\citet{finn2017deep,ebert2018visual} use a form of
MPC in the planning for their Visual foresight robotic manipulation
system (that we have seen in a previous section). The MPC part uses a model that
generates the corresponding sequence of future frames based on an
image to select the least-cost
sequence of actions.

\newcommand{\approximator}[1]{\widetilde{#1}}  
\newcommand{\MDPTransition}[0]{T}
\newcommand{\MDPAction}{\vec{a}}				
\newcommand{\MDPState}{\vec{s}} 				
\newcommand{\horizon}[0]{T}
\newcommand{\traj}[0]{\tau}
\newcommand{\MDPreward}{r}

\begin{algorithm}[t]
\small
\begin{algorithmic}
\floatname{algorithm}{Procedure}
\renewcommand{\algorithmicrequire}{\textbf{Input:}}
\renewcommand{\algorithmicensure}{\textbf{Output:}}
 \State{Initialize data $D$ with a random controller for one trial}
 \For {Trial $k = 1$ to $K$}
 \State{Train a \textit{PE} dynamics model \smash{$\approximator{\MDPTransition}$ given $D$}} 
 \For {Time $t = 0$ to TaskHorizon}
 \For {Actions sampled \smash{$\MDPAction_{t:t+\horizon}\!\sim\!\text{CEM}(\cdot)$}, 1 to NSamples}
 \State{Propagate state particles \smash{$\MDPState_{\traj}^{p}$} using \textit{TS} and \smash{$\approximator{\MDPTransition}|\{D,\MDPAction_{t:t+\horizon}\}$}}
 \State{Evaluate actions as \smash{$\sum_{\tau=t}^{t+\horizon}{\tfrac{1}{P}\sum_{p=1}^{P}{\MDPreward(\MDPState_{\tau}^{p}, \MDPAction_{\traj})}}$}}
 \State{Update \smash{$\text{CEM}(\cdot)$} distribution}
 \EndFor
 \State{Execute first action \smash{$\MDPAction^*_{t}$} (only) from optimal actions \smash{$\MDPAction^*_{t:t+\horizon}$}}
 \State{Record outcome: \smash{$D \leftarrow D \cup \{\MDPState_{t}, \MDPAction^*_{t}, \MDPState_{t+1}\}$}.}
 \EndFor
 \EndFor
\end{algorithmic}
\caption{PETS MPC~\citep{chua2018deep}}
\label{alg:pets}
\end{algorithm}

Another approach uses ensemble models for learning the transition
model, while using MPC for planning. PETS~\citep{chua2018deep} uses
probabilistic ensembles~\citep{lakshminarayanan2017simple} for
learning.  In MPC fashion  only the
first action from the CEM-optimized  sequence is used, re-planning at every
time-step (see Algorithm~\ref{alg:pets}).

MPC is a simple and effective planning method that is well-suited for
model inaccuracy, by restricting the planning horizon. MPC
is used with success in model-based reinforcement learning, with
high-variance or complex transition models. MPC has also been used with success in combination with
latent models~\citep{hafner2018learning,kaiser2019model}.

\subsubsection{End-to-End Learning and Planning}
Our third planning approach is different, it integrates planning with
learning in a fully differentiable algorithm. Let us see how this works.

Model-based reinforcement learning consists of two distinct functions:
\emph{learning} the transition model and \emph{planning} with the
model to improve the behavior policy (Figure~\ref{fig:learn}). In classical, tabular,
reinforcement learning, both learning and planning functions are
designed by hand~\citep{sutton2018introduction}. In deep reinforcement learning,
one of these functions is approximated by deep learning---the model
learning---while the planner is still hand-written.
End-to-end  learning and planning breaks this classical planning
barrier. End-to-end approaches  integrate the planning into deep
learning, using differentiable planning algorithms, extending the
backpropagation fully from reward to observation in all parts of the
model-based approach.

How can  a neural network learn to plan? 
While conceptually exciting and appealing, there are challenges to
overcome. Among them are finding suitable differentiable planning algorithms
and the  increase in  computational training complexity, since
now the planner must also be learned.

The idea of planning by gradient descent exists for some 
time, several authors explored learning approximations of state
transition  dynamics in neural
networks~\citep{kelley1960gradient,schmidhuber1990line,ilin2007efficient}. 
Neural networks are typically used to transform and filter, to learn
selection and classification tasks. A planner unrolls a state,
computes values, using selection and value aggregation, and backtracks
to try another state. Although counter-intuitive at first, these
operations are not that different from what classic neural networks are
performing. A progression of papers has published methods on how this
can be achieved.

We will start at the beginning, with convolutional neural networks
(CNN) and value iteration. We will see how the
iterations of value iteration can be implemented 
in the layers of a convolutional neural network (CNN).
Next, two variations of this method are presented, and a
way to implement planning with convolutional LSTM modules. All these
approaches implement differentiable, trainable, planning algorithms,
that can generalize to different inputs. The later methods use
elaborate schemes with
latent models so that the learning can be applied to different application domains.

Let us start to see what is possible with a CNN.
A CNN can be used to implement value iteration. This was first shown
by~\citet{tamar2016value}, who introduced value iteration networks
(VIN).
The core  idea is that value
iteration (VI, see Algorithm~\ref{lst:vi}) 
can be implemented step-by-step by a multi-layer convolutional
network: each layer does a step of lookahead. In this way VI is
implemented in a CNN. The VI iterations for
the Q-action-value-function are rolled out in
the network layers $Q$ with $A$ channels.
Through backpropagation the model learns the value
function. The aim is to learn a general model, that can
navigate in unseen environments.

VIN can be used for discrete and continuous path planning, and has
been tried in  grid world problems and natural language
tasks. VIN has achieved generalization of finding shortest paths in unseen mazes.
However, a limitation of VIN is that the number of
layers of the CNN restricts the number of planning steps, restricting
VINs  to small and low-dimensional domains. Follow-up studies
focus on making end-to-end learning and planning more generally
applicable.  \citet{schleich2019value} extend VINs by adding abstraction, and
\citet{srinivas2018universal} introduce universal planning networks, UPN,
which generalize to modified robot morphologies. Value
propagation~\citep{nardelli2018value} uses a hierarchical structure to
generalize end-to-end methods to large
problems. TreeQN~\citep{farquhar2018treeqn} incorporates a recursive
tree structure in the network, modeling the different functions of an
MDP  explicitly. TreeQN is applied to Sokoban and nine Atari games.

A further step is to model more complex planning algorithms, such as
Monte Carlo Tree Search (MCTS), a  successful planning
algorithm~\citep{coulom2006efficient,browne2012survey}. This
has been achieved to a certain extent by~\citet{guez2018learning} who
implement many elements of MCTS in MCTSnets
and~\citep{guez2019investigation}. 
In this method planning is learned with a general recurrent architecture
consisting of LSTMs and a convolutional
network~\citep{schmidhuber1990making} in the form of a stack of
ConvLSTM modules~\citep{xingjian2015convolutional}.
The architecture was used on Sokoban and
boxworld~\citep{zambaldi2018relational}, and was able to perform full
planning steps. Future work should investigate how to achieve sample-efficiency with
this architecture.

The question whether  model-based planning can be learned by a neural network has been
studied by~\citep{pascanu2017learning}, who showed  that imagination-based
planning steps can indeed be learned for a small game with an LSTM.
Related to this, imagination-augmented agents (I2A) has been designed as a fully end-to-end
differentiable architecture for model-based
imagination and  model-free reinforcement learning~\citep{racaniere2017imagination}. It consists of an
LSTM-based encoder~\citep{chiappa2017recurrent,buesing2018learning}, a ConvLSTM rollout module, and a standard
CNN-based model-free path. The policy improvement algorithm is A3C. 
\citet{racaniere2017imagination} report that on Sokoban and Pacman I2A performs better than
model-free learning and MCTS.
I2A has been specifically designed to handle model imperfections
well and  uses
a manager or meta-controller to choose between 
rolling out actions in the environment or by imagination~\citep{hamrick2017metacontrol}.


In VIN there is a tight connection between the  network architecture
and the application structure.  One way to remedy this restriction is
with a latent model, such as the ones that were
discussed earlier. One of the first attempts is the
Predictron~\citep{silver2017predictron}, where the familiar four
elements appear: a representation model, a transition model, a reward model, and a
value model. The goal of the latent model is to perform value
prediction (not state prediction), including being able to encode
special events such as ``staying alive'' or ``reaching the next
room.'' Predictron performs limited-horizon rollouts, and has been
applied to procedurally generated mazes.

One of the main success stories of model-based reinforcement learning
is AlphaZero~\citep{silver2017mastering,silver2018general}. AlphaZero
combines planning and learning in a highly successful way. Inspired by
this approach, a fully differentiable version of this architecture has
been introduced by~\citet{schrittwieser2020mastering} which is named
MuZero. This system is able to learn the rules and to learn
to play games as different
as Atari, chess, and Go, purely from the environment, with end-to-end
learning and planning. The MuZero architecture is based on Predictron,
with an abstract model consisting of a representation, transition,
reward, and prediction function (policy and value). For planning,
MuZero uses an explicitly coded version of MCTS that uses policy and
value input form the network~\citep{rosin2011multi}, but  that is executed
separately from  the network.

End-to-end planning and learning has shown impressive results, but
there are still open questions concerning the applicability to
different applications, and especially the scalability to larger problems.

\subsection{Applications}
After we have discussed in some depth the learning and planning
methods, we must look at the third element of the taxonomy: the
applications. We will see that the type of application plays an
important role in the success of the learning and planning methods.

Model-based reinforcement learning is applied to sequential decision
problems. Which types of sequential decision problems can we
distinguish? Two main application areas are robotics and games.  The
actions in robotics  are continuous, and the environment is
non-deterministic. The actions in games
are typically discrete and the environment is often deterministic. We
will describe four application areas:
\begin{enumerate}
\item continuous action space
  \begin{enumerate}
  \item small tasks
  \item large tasks
  \end{enumerate}
\item discrete action space
  \begin{enumerate}
  \item low-dimensional input
  \item high-dimensional input
  \end{enumerate}
\end{enumerate}

\subsubsection{Continuous Actions}
Sequential decision problems are well-suited to model robotic actions,
such as how to move the joints in a robotic arm to pour a cup of tea,
how to move the joints of a humanoid figure to stand up when lying
down, and how to develop gaits of a four-legged  animal. The
action space of such problems is continuous since the angles over
which robotic joints move  span a continuous range of
values. Furthermore, the environment in which 
robots operate  mimics the real world, and is 
non-deterministic. Things move, objects do not always respond in a
predictable fashion, and unexpected situations arise.

In reinforcement learning, where agent algorithms are trained 
by the feedback on their many actions, working with real robots would get
prohibitively expensive due to wear. Most reinforcement learning
systems use physics simulations such as offered by
MuJoCo~\citep{todorov2012mujoco}. MuJoCo allows the creation of
experiments that provide environments for an agent. Tasks can range
form small to  large.

\begin{figure}[t]
 \begin{center}
 \includegraphics[width=\textwidth]{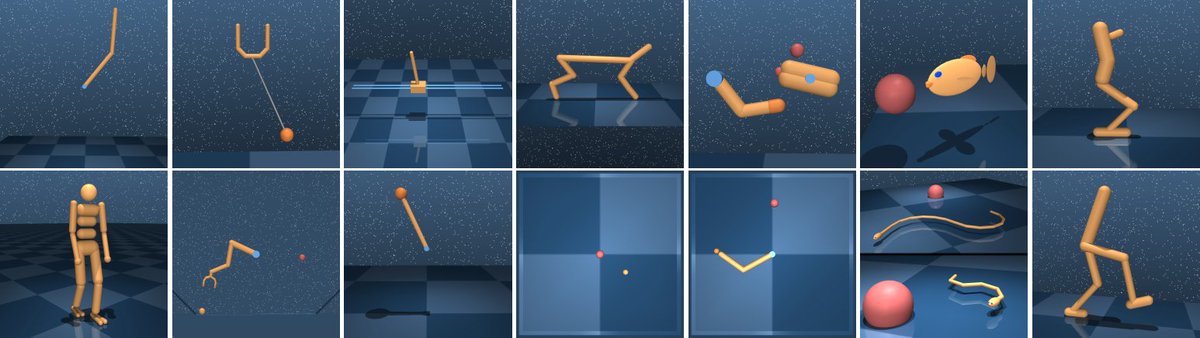}
 \caption[DeepMind Control Suite]{DeepMind Control Suite. Top: Acrobot, Ball-in-cup, Cart-pole, Cheetah, Finger, Fish, Hopper. Bottom: Humanoid, Manipulator, Pendulum, Point-mass, Reacher, Swimmer (6 and 15 links), Walker}\label{fig:dcs}
 \end{center}
\end{figure}

\begin{figure}[t]
  \begin{center}
    \includegraphics[width=8cm]{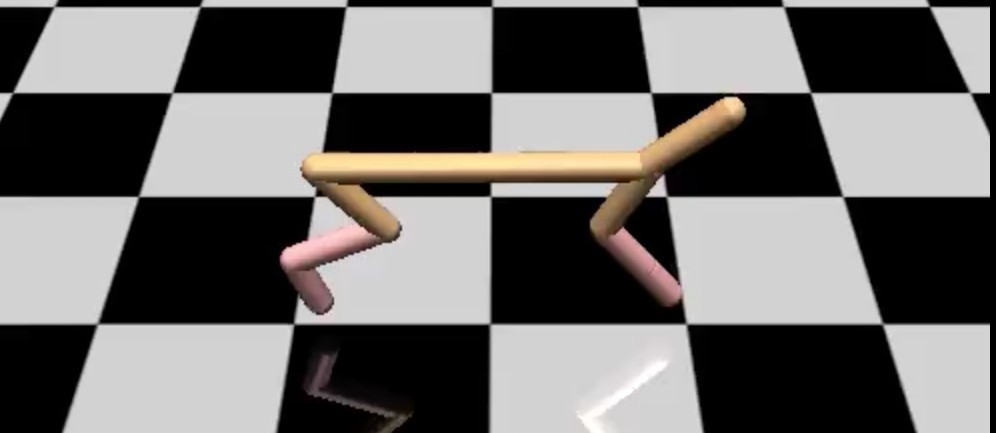}
  \end{center}
  \caption{Half-Cheetah}\label{fig:cheetah}
\end{figure}

\subsubsection*{Small}
MuJoCo tasks differ in difficulty, depending on how many
joints or degrees of freedom are modeled, and which task is being
learned. Figure~\ref{fig:dcs} shows 
some of the tasks that have been modeled in MuJoCo as part of the
DeepMind Control Suite~\citep{tassa2018deepmind}. Some of the small
tasks are ball-in-cup and reacher. The iterative quadratic non-linear
optimization method
iLQG~\citep{tassa2012synthesis} is able to teach a humanoid to stand
up, and Guided policy search~\citep{levine2013guided} and Stochastic
value gradients~\citep{heess2015learning} can learn tasks such as
swimmer, reacher, half-cheetah (Figure~\ref{fig:cheetah}) and walker. Also ensemble methods
such as PETS, MVE, and meta ensembles achieve good results on these
applications
~\citep{gu2016continuous,chua2018deep,feinberg2018model,clavera2018model,janner2019trust}.

\subsubsection*{Large}
MuJoCo has enabled progress in model-based reinforcement learning in small
continuous tasks, and also in larger tasks, such as  how to develop
gaits of a four-legged robotic animal, or how to scale an obstacle
course. PlaNet~\citep{hafner2018learning}, Dreamer~\citep{hafner2019dream} and
MBPO~\citep{janner2019trust} achieve good results on more complicated
MuJoCo tasks using latent models to reduce the dimensionality.

\subsubsection{Discrete Actions}
There is a long tradition in reinforcement learning  to see if we can
teach a computer to play complicated games and puzzles~\citep{plaat2020learning}. 
Games and puzzles are often played on a board with discrete
squares. Actions in such games are discrete, a move to square e3 is
not a move to square e4. The environments are also deterministic, we
assume that pieces do not move by itself.

Most games that are used in deep model-based reinforcement learning papers 
fall into this category. More complex games, such as partial
information (card games such as poker~\citep{brown2019superhuman}) or
games with multiple actors   (real-time strategy video games such as
StarCraft~\citep{vinyals2019grandmaster,ontanon2013survey,wong2021survey}) are not used in the 
approaches that we survey here.

\begin{figure}[t]
  \begin{center}
    \begin{tabular}{cc}
      \includegraphics[width=5cm]{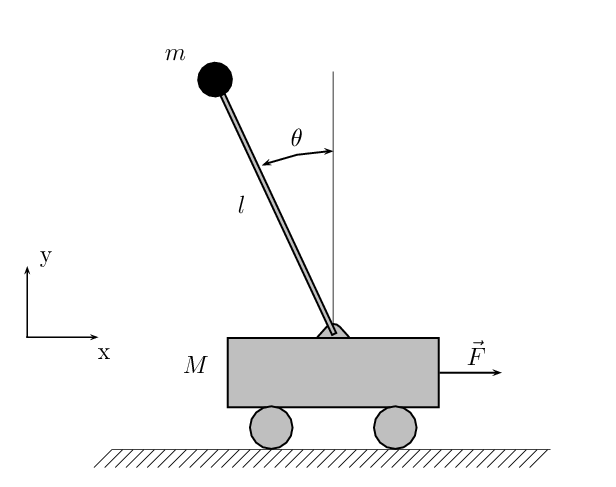}
      &
      \includegraphics[width=5cm]{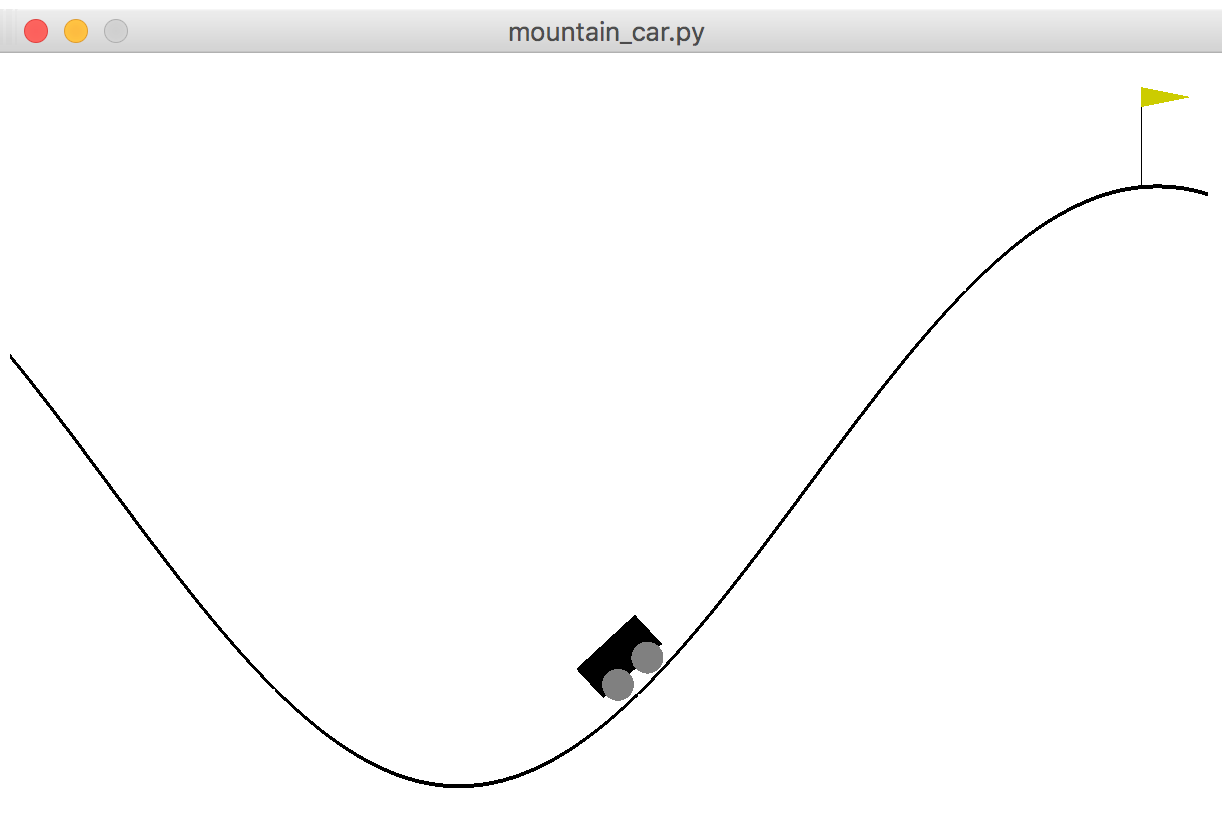}
    \end{tabular}
 \caption{Cart Pole and Mountain Car}\label{fig:cartpole4}
 \end{center}
\end{figure}

\subsubsection*{Low-Dimensional}
Among low-dimensional applications that are used in model-based
reinforcement learning are simple pendulum problems, Cartpole and
Mountain car, where the challenge is to reverse engineer the laws of impulse and
gravity  (Figure~\ref{fig:cartpole4}). The action space consists of
two discrete actions,
push left or push right, the environment is continuous and
deterministic. PILCO~\citep{deisenroth2011pilco} achieves good results
with Gaussian process modeling and gradient based planning on the
pendulum task.
\begin{figure}[t]
\begin{center}
\includegraphics[width=5cm]{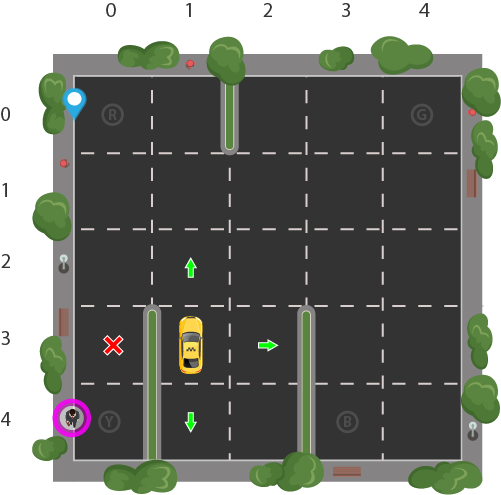}
\caption{Taxi in a Grid world~\citep{dietterich1998maxq,learn}}\label{fig:grid}
\end{center}
\end{figure}

Perhaps the most frequently used low-dimensional application area is grid-world,
where various navigation tasks are tested (Figure~\ref{fig:grid}). VIN~\citep{tamar2016value},
VProp~\citep{nardelli2018value} and the
Predictron~\citep{silver2017predictron} that use maze navigation to
test their approaches to integrating
end-to-end  learning and planning. Grid worlds and mazes can be designed
and scaled in different forms and sizes, making them well suited for
testing new ideas.

\begin{figure}[t]
  \begin{center}
    \includegraphics[width=7cm]{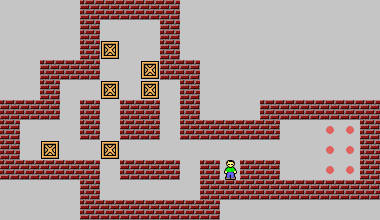}
  \end{center}
  \caption{Sokoban Puzzle~\citep{chao2013}}\label{fig:sokoban}
\end{figure}

Other low-dimensional games are board games such as chess and Go. These games are
low-dimensional (their input has few atttributes, in contrast to a
mega-pixel image) but they nevertheless have a large state
space. Finding good policies for chess and Go was one of the
most challenging feats in reinforcement
learning~\citep{campbell2002deep,silver2016mastering,plaat2020learning}. Block puzzles
such as Sokoban (Figure~\ref{fig:sokoban}), are also often used to
test reinforcement learning methods. 
Sokoban is a
block-pushing puzzle that derives much of its complexity from the
facft that the agent can push a box, but cannot pull (undo) a mistake,
giving rise to many dead-ends that are hard to detect. It has been used by
I2A~\citep{racaniere2017imagination} and MCTS network
planning~\citep{guez2019investigation} approaches that implement
planning by unrolling  steps  within a neural network.

\subsubsection*{High-dimensional}
\begin{figure}[t]
\begin{center}
\includegraphics[width=7cm]{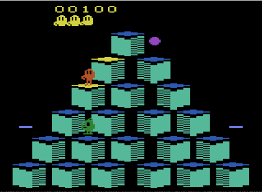}
\caption{Q*bert, Example Game from the Arcade Learning Environment}\label{fig:ale}
\end{center}
\end{figure}

Most recent success in model-free reinforcement learning has been
achieved in high-dimensional problems, such as the Arcade Learning
Environment~\citep{bellemare2013arcade}, see for
example~\citep{mnih2015human,hessel2017rainbow}. Atari games were
popular video games in the 1980s in game arcades. Figure~\ref{fig:ale}
shows a screenshot of Q*bert, a typical  Atari arcade game.

Deep learning methods are well suited to process high-dimensional
inputs. The challenge for model-based methods is to learn accurate
models with a  low number of observations. Latent model
approaches have been tried for Atari with some success~\citep{oh2015action,oh2017value,kaiser2019model,hafner2020mastering}. The
MuZero approach  was even able to
learn the rules of very different games: Atari, Go,  shogi
(Japanese chess) and chess~\citep{schrittwieser2020mastering}.

\section{Discussion and Outlook}\label{sec:dis}
We have now discussed in depth our taxonomy of application, learning
method, and planning method. We have seen  different innovative
approaches, all aiming to achieve similar goals. Let us discuss how
well they succeed.

Model-based reinforcement learning  promises high accuracy and low sample
complexity. Sutton's work on imagination, where
a transition model is created
with environment samples that are then used to create extra ``imagined'' samples
for the policy for free, clearly suggests this aspect of model-based reinforcement
learning. The  transition model acts as a multiplier on the
amount of information that is used from each environment sample, as
the agent builds up its own  model of the environment.

Another, and perhaps  more important aspect, is generalization
performance. Model-based reinforcement learning builds a dynamics model of the
environment. This model can be used multiple times, not only for the
same problem instance, but also for new problem
instances, and for variations. By learning the state-to-state
transition model and the reward model, model-based reinforcement 
learning   captures the
essence of a domain, where
model-free methods only learn best response actions. Model-based reinforcement
learning may thus be better suited for solving transfer learning problems,
and for solving long sequential decision making problems.
It is the difference  between learning how to respond to certain
actions  of a
difficult boss, and \emph{knowing} the boss.

\subsection{The Challenge}
Classical tabular approaches and Gaussian process approaches have
been quite succesful in achieving low sample complexity for problems
of small
complexity~\citep{sutton2018introduction,deisenroth2013survey,kober2013reinforcement}.
However,
the topic of the current survey is  \emph{deep} models, for large, high 
dimensional, problems with complex, non-linear, and discontinuous
functions. These application domains pose a problem for classical  approaches.

The main challenge that the model-based reinforcement learning
algorithms in this survey  address is the following.
For high-dimensional tasks the curse of dimensionality 
causes data to be sparse and variance to be high. Deep  methods tend to
overfit on small sample sizes, and  model-free methods
use   many millions of environment
samples. For high-dimensional problems, the accuracy of transition models
is under pressure. Model-based methods that use poor 
models make poor planning predictions far into the
future~\citep{talvitie2015agnostic}.

The challenge is therefore to learn deep, high-dimensional transition functions from limited
data that are accurate---or that  account for model uncertainty---and plan over these
models to achieve  policy and value functions that are as accurate or
better than model-free methods.


\subsection{Taxonomy}
In discussing our survey, we first summarize the taxonomy.
Model-based methods use two main algorithms: (1) an algorithm that is
used to learn the model, 
and (2) an algorithm that is used to plan the policy based on the
model (Figure~\ref{fig:learn}).
Our taxonomy is based on application characteristics and algorithm
characteristics---Table~\ref{tab:tax2}: application, learning, planning. First of all, we
distinguish continuous versus discrete action spaces (robotics versus games).
Second, we look at the algorithms. For the model-learning algorithms
we look at latent models, uncertainty modeling and ensembles
for continuous action spaces, and for discrete action spaces  we
look at latent and end-to-end learning and planning.
For the planning algorithms
we look at trajectory rollouts and model-predictive control
for continuous action spaces, and  for discrete action spaces we
look at end-to-end learning and planning and trajectory rollouts.

%
\begin{table}[t]
  \begin{center}\footnotesize
    \begin{tabular}{lll}
      {\bf Application}&{\bf Learning}&{\bf Planning}\\
      \hline\hline
      {Discrete}   &  CNN/LSTM             & End-to-end learning and planning\\
                   &  Latent models        & Trajectory rollouts\\ \hline
      {Continuous} & Latent models         & Trajectory rollouts\\
                   & Uncertainty  modeling & Model-predictive control\\
                   & Ensemble models       & \\
      \hline
      
    \end{tabular}
    \caption{Taxonomy:
      {\em Application, Learning, Planning}}\label{tab:tax2}
  \end{center}
\end{table}

\subsection{Outcome}
We can now consider the question if finding methods to achieve
high-accuracy models with low sample complexity has been solved.
This is the central research question that many researchers have
worked on.
Unfortunately, authors have used different
tasks within benchmark suites---or even different  benchmarks---making comparisons
between different publications challenging. A study
by~\citet{wang2019benchmarking} reimplemented many  methods
for  continuous problems to perform a fair comparison. They found that ensemble methods and
model-predictive control indeed achieve good  
results on MuJoCo tasks, and  do so in significantly fewer time
steps than model-free methods.  Typcially model-based used 200k time steps versus 1
million for model-free. However, they also found that although the
sample complexity is less, the 
wall-clock time may be  different, with model-free methods such as
PPO~\citep{schulman2017proximal} and
SAC~\citep{haarnoja2018soft,haarnoja2019soft} being much faster for
some problems. The accuracy for the policy 
varies greatly for different problems, as does the sensitivity to different
hyperparameter values; i.e., results are brittle.

Taking  these caveats into consideration, in general we conclude that the papers that
we survey report that, for high-dimensional problems, 
model-based methods do indeed approach an accuracy
as high as model-free baselines, with substantially fewer environment samples.
Therefore we conclude that the methods that we survey overcome the difficulties
posed by overfitting. Indeed, we have seen that new classes of applications
have become possible,
both in continuous action spaces---learning to perform complex robotic
behaviors---and in discrete action spaces---learning the rules of
Atari, chess, shogi, and Go.




\subsection{Further Challenges}
Although in some important applications high accuracy at lower sample complexity has been
achieved, and although quite a few results are impressive, challenges
remain. To start,  the algorithms that have been developed are quite complex. Latent models,
end-to-end learning and planning, and uncertainty modeling, are complex algorithms, that
require  effort to understand, and even more to implement
correctly in new applications. In addition, the
learned transition models are used in single problems only. Few results are reported where
they are used in a transfer learning or meta learning setting~\citep{brazdil2008metalearning,hospedales2020meta,huisman2021survey}, with the
exception of~\citet{sekar2020planning}. 

Furthermore, reproducibility is a challenge due to the use of
different benchmarks. Also, high sensitivity to differences in hyperparameter values
leads to  brittleness in results, making reprodicibility 
difficult. Finally, we note that 
continuous problems that are solved appear to be of lower
dimensionality than some  discrete problems.



\subsection{Research Agenda}

The good results  are promising for future work.
Based on the results that have been achieved, and the challenges that remain,
we come to the following research agenda.

We note that different approaches were developed for different applications.
Reproducibility of results is a challenge, different hyperparameters
can have a
large influence on performance. Furthermore, there are many different
benchmarks in use in the field, and the complexity of some continuous benchmarks
is less than for discrete. Reproducibility and benchmarking are the \emph{first}
item on our research agenda.

We also note that end-to-end learning and planning is a complex algorithm, that does
not work on all applications, and that requires large computational
effort. Applying end-to-end to large problems is still a challenge.
As \emph{second} item on our research agenda is to develop end-to-end learning and planning
further, to become more efficient, and to be applicable to more and different applications.

We further note that latent models and world models are also complex. Different
approaches use difference types of modules, sometimes consisting of submodules.
As \emph{third} item on our research agenda is  to integrate latent models
with end-to-end learning and planning,  as \emph{fourth} item, we would
like to simplify latent models and standardize them, if possible for
different applications, and, as \emph{fifth} item, we wish to apply latent models to
higher-dimensional continuous problems.

Finally, we would like to enter on our research agenda meta and
transfer learning experiments for model-based reinforcement learning,
as \emph{sixth} item.

In summary, to improve the accuracy and applicability of model-based
methods we suggest to work on the following:
\begin{enumerate}
\item Improve reproducibility of model-based reinforcement learning,
  standardize benchmarking, and improve robustness (hyperparameters)
\item Improve efficiency and of integrated end-to-end learning and
  planning; improve   applicability to more and larger applications
\item Integrate latent models and end-to-end learning and planning
\item Simplify the latent model architecture across different
  applications
\item Apply latent models to more higher-dimensional continuous problems
\item Use model-based reinforcement learning transition models for meta and transfer
  learning
\end{enumerate}

\section{Conclusion}\label{sec:con}
Deep learning has revolutionized  reinforcement
learning. The new methods allow us to approach more complicated
problems than before. Control and decision making tasks involving
high dimensional visual input have come within reach.

Model-based methods offer the advantage of lower sample
complexity than model-free methods, because agents learn  their own
transition model of the environment.
However, traditional methods, such as Gaussian processes, that work well on
moderately complex problems with few samples, do not perform
well on high-dimensional problems. High-capacity models may have high
sample complexity to create high-accuracy models, and finding methods
that generalize well with low 
sample complexity has been difficult.

In the last five years many new methods have been devised,
and great success has been achieved  in model-based deep
reinforcement learning. This survey summarizes the main ideas of
recent papers in a taxonomy based on applications and algorithms.  Latent models condense
complex problems into compact latent representations that are easier
to learn, improving the accuracy of the model; limited horizon
planning reduces the impact of low-accuracy models; 
end-to-end  methods have been devised to integrate learning and
planning in one fully differentiable approach. 
The Arcade Learning Environment has been one of the main benchmarks in
model-free reinforcement learning, starting off the recent interest in
the field with the work by~\citet{mnih2013playing,mnih2015human}. The
high-dimensionality of Atari video input has long been problematic for
model-based reinforcement learning. Latent models were instrumental in
reducing the dimensionality of Atari, producing the first successes
for model-based approaches on this major benchmark.
Despite this success,  challenges remain.
In the discussion we mentioned  open problems for each of the
approaches, where we expect worthwhile  future work to
occur.
Impressive results have been  
reported; future work can be expected in transfer learning with 
latent models, and the interplay of   latent
models, in combination  with end-to-end learning of larger problems.
Benchmarks in the field have also had to keep up.
Benchmarks have progressed from single-agent grid worlds to
high-dimensional games and  
complicated camera-arm  manipulation tasks.
Reproducibility and benchmarking studies are of great importance for
real progress. In real-time strategy games model-based methods are 
being combined with multi-agent, hierarchical  and evolutionary
approaches, allowing the study of collaboration, competation and
negotiation.

Model-based deep reinforcement learning is a vibrant field of
AI with a long history before deep learning. The field is blessed with
a high degree of activity, 
an open culture, clear benchmarks,  shared
code-bases~\citep{bellemare2013arcade,brockman2016openai,vinyals2017starcraft,tassa2018deepmind}
and a quick turnaround of 
ideas. We hope that this survey will contribute to 
the low barrier of entry.

\section*{Acknowledgments}
We thank the members of the Leiden Reinforcement
Learning Group, and especially Thomas Moerland and Mike Huisman,
for many
discussions and insights.

\bibliographystyle{plainnat}
\bibliography{\string~/Dropbox/BibTex/plaat}

\begin{thebibliography}{120}
\providecommand{\natexlab}[1]{#1}
\providecommand{\url}[1]{\texttt{#1}}
\expandafter\ifx\csname urlstyle\endcsname\relax
  \providecommand{\doi}[1]{doi: #1}\else
  \providecommand{\doi}{doi: \begingroup \urlstyle{rm}\Url}\fi

\bibitem[Abbeel et~al.(2007)Abbeel, Coates, Quigley, and
  Ng]{abbeel2007application}
Pieter Abbeel, Adam Coates, Morgan Quigley, and Andrew~Y Ng.
\newblock An application of reinforcement learning to aerobatic helicopter
  flight.
\newblock In \emph{Advances in Neural Information Processing Systems}, pages
  1--8, 2007.

\bibitem[Alpaydin(2020)]{alpaydin2020introduction}
Ethem Alpaydin.
\newblock \emph{Introduction to machine learning, Third edition}.
\newblock MIT Press, 2020.

\bibitem[Anthony et~al.(2017)Anthony, Tian, and Barber]{anthony2017thinking}
Thomas Anthony, Zheng Tian, and David Barber.
\newblock Thinking fast and slow with deep learning and tree search.
\newblock In \emph{Advances in Neural Information Processing Systems}, pages
  5360--5370, 2017.

\bibitem[Bellemare et~al.(2013)Bellemare, Naddaf, Veness, and
  Bowling]{bellemare2013arcade}
Marc~G Bellemare, Yavar Naddaf, Joel Veness, and Michael Bowling.
\newblock The {Arcade Learning Environment}: An evaluation platform for general
  agents.
\newblock \emph{Journal of Artificial Intelligence Research}, 47:\penalty0
  253--279, 2013.

\bibitem[Bellman(1957, 2013)]{bellman1957dynamic}
Richard Bellman.
\newblock \emph{Dynamic programming}.
\newblock Courier Corporation, 1957, 2013.

\bibitem[Bishop(2006)]{bishop2006pattern}
Christopher~M Bishop.
\newblock \emph{Pattern recognition and machine learning}.
\newblock Information science and statistics. Springer Verlag, Heidelberg,
  2006.

\bibitem[Botev et~al.(2013)Botev, Kroese, Rubinstein, and
  L'Ecuyer]{botev2013cross}
Zdravko~I Botev, Dirk~P Kroese, Reuven~Y Rubinstein, and Pierre L'Ecuyer.
\newblock The cross-entropy method for optimization.
\newblock In \emph{Handbook of statistics}, volume~31, pages 35--59. Elsevier,
  2013.

\bibitem[Brazdil et~al.(2008)Brazdil, Carrier, Soares, and
  Vilalta]{brazdil2008metalearning}
Pavel Brazdil, Christophe~Giraud Carrier, Carlos Soares, and Ricardo Vilalta.
\newblock \emph{Metalearning: Applications to data mining}.
\newblock Springer Science \& Business Media, 2008.

\bibitem[Brockman et~al.(2016)Brockman, Cheung, Pettersson, Schneider,
  Schulman, Tang, and Zaremba]{brockman2016openai}
Greg Brockman, Vicki Cheung, Ludwig Pettersson, Jonas Schneider, John Schulman,
  Jie Tang, and Wojciech Zaremba.
\newblock {OpenAI} {G}ym.
\newblock \emph{arXiv preprint arXiv:1606.01540}, 2016.

\bibitem[Brown and Sandholm(2019)]{brown2019superhuman}
Noam Brown and Tuomas Sandholm.
\newblock Superhuman {AI} for multiplayer poker.
\newblock \emph{Science}, 365\penalty0 (6456):\penalty0 885--890, 2019.

\bibitem[Browne et~al.(2012)Browne, Powley, Whitehouse, Lucas, Cowling,
  Rohlfshagen, Tavener, Perez, Samothrakis, and Colton]{browne2012survey}
Cameron~B Browne, Edward Powley, Daniel Whitehouse, Simon~M Lucas, Peter~I
  Cowling, Philipp Rohlfshagen, Stephen Tavener, Diego Perez, Spyridon
  Samothrakis, and Simon Colton.
\newblock A survey of {Monte} {Carlo} {Tree} {Search} methods.
\newblock \emph{IEEE Transactions on Computational Intelligence and AI in
  Games}, 4\penalty0 (1):\penalty0 1--43, 2012.

\bibitem[Buesing et~al.(2018)Buesing, Weber, Racaniere, Eslami, Rezende,
  Reichert, Viola, Besse, Gregor, Hassabis, et~al.]{buesing2018learning}
Lars Buesing, Theophane Weber, S{\'e}bastien Racaniere, SM~Eslami, Danilo
  Rezende, David~P Reichert, Fabio Viola, Frederic Besse, Karol Gregor, Demis
  Hassabis, et~al.
\newblock Learning and querying fast generative models for reinforcement
  learning.
\newblock \emph{arXiv preprint arXiv:1802.03006}, 2018.

\bibitem[{\c{C}}al{\i}{\c{s}}{\i}r and
  Pehlivano{\u{g}}lu(2019)]{ccalicsir2019model}
Sinan {\c{C}}al{\i}{\c{s}}{\i}r and Meltem~Kurt Pehlivano{\u{g}}lu.
\newblock Model-free reinforcement learning algorithms: A survey.
\newblock In \emph{2019 27th Signal Processing and Communications Applications
  Conference (SIU)}, pages 1--4, 2019.

\bibitem[Campbell et~al.(2002)Campbell, Hoane~Jr, and Hsu]{campbell2002deep}
Murray Campbell, A~Joseph Hoane~Jr, and Feng-hsiung Hsu.
\newblock Deep {Blue}.
\newblock \emph{Artificial Intelligence}, 134\penalty0 (1-2):\penalty0 57--83,
  2002.

\bibitem[Chao(2013)]{chao2013}
Yang Chao.
\newblock Sokoban.org, 2013.

\bibitem[Chiappa et~al.(2017)Chiappa, Racaniere, Wierstra, and
  Mohamed]{chiappa2017recurrent}
Silvia Chiappa, S{\'e}bastien Racaniere, Daan Wierstra, and Shakir Mohamed.
\newblock Recurrent environment simulators.
\newblock \emph{arXiv preprint arXiv:1704.02254}, 2017.

\bibitem[Chua et~al.(2018)Chua, Calandra, McAllister, and Levine]{chua2018deep}
Kurtland Chua, Roberto Calandra, Rowan McAllister, and Sergey Levine.
\newblock Deep reinforcement learning in a handful of trials using
  probabilistic dynamics models.
\newblock In \emph{Advances in Neural Information Processing Systems}, pages
  4754--4765, 2018.

\bibitem[Clavera et~al.(2018)Clavera, Rothfuss, Schulman, Fujita, Asfour, and
  Abbeel]{clavera2018model}
Ignasi Clavera, Jonas Rothfuss, John Schulman, Yasuhiro Fujita, Tamim Asfour,
  and Pieter Abbeel.
\newblock Model-based reinforcement learning via meta-policy optimization.
\newblock \emph{arXiv preprint arXiv:1809.05214}, 2018.

\bibitem[Coulom(2006)]{coulom2006efficient}
R{\'e}mi Coulom.
\newblock Efficient selectivity and backup operators in {Monte-Carlo} {Tree
  Search}.
\newblock In \emph{International Conference on Computers and Games}, pages
  72--83. Springer, 2006.

\bibitem[Deisenroth and Rasmussen(2011)]{deisenroth2011pilco}
Marc Deisenroth and Carl~E Rasmussen.
\newblock Pilco: A model-based and data-efficient approach to policy search.
\newblock In \emph{Proceedings of the 28th International Conference on Machine
  Learning (ICML-11)}, pages 465--472, 2011.

\bibitem[Deisenroth et~al.(2013)Deisenroth, Neumann, and
  Peters]{deisenroth2013survey}
Marc~Peter Deisenroth, Gerhard Neumann, and Jan Peters.
\newblock A survey on policy search for robotics.
\newblock In \emph{rFoundations and Trends in Robotics 2}, pages 1--142. Now
  publishers, 2013.

\bibitem[Dietterich(1998)]{dietterich1998maxq}
Thomas~G Dietterich.
\newblock The maxq method for hierarchical reinforcement learning.
\newblock In \emph{ICML}, volume~98, pages 118--126. Citeseer, 1998.

\bibitem[Doerr et~al.(2018)Doerr, Daniel, Schiegg, Nguyen-Tuong, Schaal,
  Toussaint, and Trimpe]{doerr2018probabilistic}
Andreas Doerr, Christian Daniel, Martin Schiegg, Duy Nguyen-Tuong, Stefan
  Schaal, Marc Toussaint, and Sebastian Trimpe.
\newblock Probabilistic recurrent state-space models.
\newblock \emph{arXiv preprint arXiv:1801.10395}, 2018.

\bibitem[Duan et~al.(2016)Duan, Schulman, Chen, Bartlett, Sutskever, and
  Abbeel]{duan2016rl}
Yan Duan, John Schulman, Xi~Chen, Peter~L Bartlett, Ilya Sutskever, and Pieter
  Abbeel.
\newblock {RL}${}^{2}$: Fast reinforcement learning via slow reinforcement
  learning.
\newblock \emph{arXiv preprint arXiv:1611.02779}, 2016.

\bibitem[Ebert et~al.(2018)Ebert, Finn, Dasari, Xie, Lee, and
  Levine]{ebert2018visual}
Frederik Ebert, Chelsea Finn, Sudeep Dasari, Annie Xie, Alex Lee, and Sergey
  Levine.
\newblock Visual foresight: Model-based deep reinforcement learning for
  vision-based robotic control.
\newblock \emph{arXiv preprint arXiv:1812.00568}, 2018.

\bibitem[Farquhar et~al.(2018)Farquhar, Rockt{\"a}schel, Igl, and
  Whiteson]{farquhar2018treeqn}
Gregory Farquhar, Tim Rockt{\"a}schel, Maximilian Igl, and SA~Whiteson.
\newblock {TreeQN and ATreeC}: Differentiable tree planning for deep
  reinforcement learning.
\newblock International Conference on Learning Representations, 2018.

\bibitem[Feinberg et~al.(2018)Feinberg, Wan, Stoica, Jordan, Gonzalez, and
  Levine]{feinberg2018model}
Vladimir Feinberg, Alvin Wan, Ion Stoica, Michael~I Jordan, Joseph~E Gonzalez,
  and Sergey Levine.
\newblock Model-based value estimation for efficient model-free reinforcement
  learning.
\newblock \emph{arXiv preprint arXiv:1803.00101}, 2018.

\bibitem[Finn and Levine(2017)]{finn2017deep}
Chelsea Finn and Sergey Levine.
\newblock Deep visual foresight for planning robot motion.
\newblock In \emph{2017 IEEE International Conference on Robotics and
  Automation (ICRA)}, pages 2786--2793. IEEE, 2017.

\bibitem[Finn et~al.(2017)Finn, Abbeel, and Levine]{finn2017model}
Chelsea Finn, Pieter Abbeel, and Sergey Levine.
\newblock {Model-Agnostic Meta-Learning} for fast adaptation of deep networks.
\newblock \emph{arXiv preprint arXiv:1703.03400}, 2017.

\bibitem[Garcia et~al.(1989)Garcia, Prett, and Morari]{garcia1989model}
Carlos~E Garcia, David~M Prett, and Manfred Morari.
\newblock Model predictive control: Theory and practice---a survey.
\newblock \emph{Automatica}, 25\penalty0 (3):\penalty0 335--348, 1989.

\bibitem[Goodfellow et~al.(2014)Goodfellow, Pouget-Abadie, Mirza, Xu,
  Warde-Farley, Ozair, Courville, and Bengio]{goodfellow2014generative}
Ian Goodfellow, Jean Pouget-Abadie, Mehdi Mirza, Bing Xu, David Warde-Farley,
  Sherjil Ozair, Aaron Courville, and Yoshua Bengio.
\newblock Generative adversarial nets.
\newblock In \emph{Advances in Neural Information Processing Systems}, pages
  2672--2680, 2014.

\bibitem[Goodfellow et~al.(2016)Goodfellow, Bengio, and
  Courville]{goodfellow2016deep}
Ian Goodfellow, Yoshua Bengio, and Aaron Courville.
\newblock \emph{Deep learning}.
\newblock MIT Press, Cambridge, 2016.

\bibitem[Grimm et~al.(2020)Grimm, Barreto, Singh, and Silver]{grimm2020value}
Christopher Grimm, Andr{\'e} Barreto, Satinder Singh, and David Silver.
\newblock The value equivalence principle for model-based reinforcement
  learning.
\newblock \emph{arXiv preprint arXiv:2011.03506}, 2020.

\bibitem[Gu et~al.(2016)Gu, Lillicrap, Sutskever, and Levine]{gu2016continuous}
Shixiang Gu, Timothy Lillicrap, Ilya Sutskever, and Sergey Levine.
\newblock Continuous deep {Q}-learning with model-based acceleration.
\newblock In \emph{International Conference on Machine Learning}, pages
  2829--2838, 2016.

\bibitem[Guez et~al.(2018)Guez, Weber, Antonoglou, Simonyan, Vinyals, Wierstra,
  Munos, and Silver]{guez2018learning}
Arthur Guez, Th{\'e}ophane Weber, Ioannis Antonoglou, Karen Simonyan, Oriol
  Vinyals, Daan Wierstra, R{\'e}mi Munos, and David Silver.
\newblock Learning to search with {MCTSnets}.
\newblock \emph{arXiv preprint arXiv:1802.04697}, 2018.

\bibitem[Guez et~al.(2019)Guez, Mirza, Gregor, Kabra, Racani{\`e}re, Weber,
  Raposo, Santoro, Orseau, Eccles, et~al.]{guez2019investigation}
Arthur Guez, Mehdi Mirza, Karol Gregor, Rishabh Kabra, S{\'e}bastien
  Racani{\`e}re, Th{\'e}ophane Weber, David Raposo, Adam Santoro, Laurent
  Orseau, Tom Eccles, et~al.
\newblock An investigation of model-free planning.
\newblock \emph{arXiv preprint arXiv:1901.03559}, 2019.

\bibitem[Ha and Schmidhuber(2018{\natexlab{a}})]{ha2018recurrent}
David Ha and J{\"u}rgen Schmidhuber.
\newblock Recurrent world models facilitate policy evolution.
\newblock In \emph{Advances in Neural Information Processing Systems}, pages
  2450--2462, 2018{\natexlab{a}}.

\bibitem[Ha and Schmidhuber(2018{\natexlab{b}})]{ha2018world}
David Ha and J{\"u}rgen Schmidhuber.
\newblock World models.
\newblock \emph{arXiv preprint arXiv:1803.10122}, 2018{\natexlab{b}}.

\bibitem[Haarnoja et~al.(2018{\natexlab{a}})Haarnoja, Zhou, Abbeel, and
  Levine]{haarnoja2018soft}
Tuomas Haarnoja, Aurick Zhou, Pieter Abbeel, and Sergey Levine.
\newblock Soft actor-critic: Off-policy maximum entropy deep reinforcement
  learning with a stochastic actor.
\newblock In \emph{International conference on machine learning}, pages
  1861--1870. PMLR, 2018{\natexlab{a}}.

\bibitem[Haarnoja et~al.(2018{\natexlab{b}})Haarnoja, Zhou, Hartikainen,
  Tucker, Ha, Tan, Kumar, Zhu, Gupta, Abbeel, et~al.]{haarnoja2019soft}
Tuomas Haarnoja, Aurick Zhou, Kristian Hartikainen, George Tucker, Sehoon Ha,
  Jie Tan, Vikash Kumar, Henry Zhu, Abhishek Gupta, Pieter Abbeel, et~al.
\newblock Soft actor-critic algorithms and applications.
\newblock \emph{arXiv preprint arXiv:1812.05905}, 2018{\natexlab{b}}.

\bibitem[Hafner et~al.(2018)Hafner, Lillicrap, Fischer, Villegas, Ha, Lee, and
  Davidson]{hafner2018learning}
Danijar Hafner, Timothy Lillicrap, Ian Fischer, Ruben Villegas, David Ha,
  Honglak Lee, and James Davidson.
\newblock Learning latent dynamics for planning from pixels.
\newblock \emph{arXiv preprint arXiv:1811.04551}, 2018.

\bibitem[Hafner et~al.(2019)Hafner, Lillicrap, Ba, and
  Norouzi]{hafner2019dream}
Danijar Hafner, Timothy Lillicrap, Jimmy Ba, and Mohammad Norouzi.
\newblock Dream to control: Learning behaviors by latent imagination.
\newblock \emph{arXiv preprint arXiv:1912.01603}, 2019.

\bibitem[Hafner et~al.(2020)Hafner, Lillicrap, Norouzi, and
  Ba]{hafner2020mastering}
Danijar Hafner, Timothy Lillicrap, Mohammad Norouzi, and Jimmy Ba.
\newblock Mastering atari with discrete world models.
\newblock \emph{arXiv preprint arXiv:2010.02193}, 2020.

\bibitem[Hamrick(2019)]{hamrick2019analogues}
Jessica~B Hamrick.
\newblock Analogues of mental simulation and imagination in deep learning.
\newblock \emph{Current Opinion in Behavioral Sciences}, 29:\penalty0 8--16,
  2019.

\bibitem[Hamrick et~al.(2017)Hamrick, Ballard, Pascanu, Vinyals, Heess, and
  Battaglia]{hamrick2017metacontrol}
Jessica~B Hamrick, Andrew~J Ballard, Razvan Pascanu, Oriol Vinyals, Nicolas
  Heess, and Peter~W Battaglia.
\newblock Metacontrol for adaptive imagination-based optimization.
\newblock \emph{arXiv preprint arXiv:1705.02670}, 2017.

\bibitem[Heess et~al.(2015)Heess, Wayne, Silver, Lillicrap, Erez, and
  Tassa]{heess2015learning}
Nicolas Heess, Gregory Wayne, David Silver, Timothy Lillicrap, Tom Erez, and
  Yuval Tassa.
\newblock Learning continuous control policies by stochastic value gradients.
\newblock In \emph{Advances in Neural Information Processing Systems}, pages
  2944--2952, 2015.

\bibitem[Hessel et~al.(2017)Hessel, Modayil, Van~Hasselt, Schaul, Ostrovski,
  Dabney, Horgan, Piot, Azar, and Silver]{hessel2017rainbow}
Matteo Hessel, Joseph Modayil, Hado Van~Hasselt, Tom Schaul, Georg Ostrovski,
  Will Dabney, Dan Horgan, Bilal Piot, Mohammad Azar, and David Silver.
\newblock Rainbow: Combining improvements in deep reinforcement learning.
\newblock \emph{arXiv preprint arXiv:1710.02298}, 2017.

\bibitem[Ho(1995)]{ho1995random}
Tin~Kam Ho.
\newblock Random decision forests.
\newblock In \emph{Proceedings of 3rd international conference on document
  analysis and recognition}, volume~1, pages 278--282. IEEE, 1995.

\bibitem[Hospedales et~al.(2020)Hospedales, Antoniou, Micaelli, and
  Storkey]{hospedales2020meta}
Timothy Hospedales, Antreas Antoniou, Paul Micaelli, and Amos Storkey.
\newblock Meta-learning in neural networks: A survey.
\newblock \emph{arXiv preprint arXiv:2004.05439}, 2020.

\bibitem[Hui(2018)]{hui2018model}
Jonathan Hui.
\newblock Model-based reinforcement learning
  \url{https://medium.com/@jonathan_hui/rl-model-based-reinforcement-learning-3c2b6f0aa323}.
\newblock Medium post, 2018.

\bibitem[Huisman et~al.(2021)Huisman, van Rijn, and Plaat]{huisman2021survey}
Mike Huisman, Jan~N. van Rijn, and Aske Plaat.
\newblock A survey of deep meta-learning.
\newblock \emph{Artificial Intelligence Review}, 2021.

\bibitem[Ilin et~al.(2007)Ilin, Kozma, and Werbos]{ilin2007efficient}
Roman Ilin, Robert Kozma, and Paul~J Werbos.
\newblock Efficient learning in cellular simultaneous recurrent neural
  networks---the case of maze navigation problem.
\newblock In \emph{2007 IEEE International Symposium on Approximate Dynamic
  Programming and Reinforcement Learning}, pages 324--329, 2007.

\bibitem[Janner et~al.(2019)Janner, Fu, Zhang, and Levine]{janner2019trust}
Michael Janner, Justin Fu, Marvin Zhang, and Sergey Levine.
\newblock When to trust your model: Model-based policy optimization.
\newblock In \emph{Advances in Neural Information Processing Systems}, pages
  12498--12509, 2019.

\bibitem[Justesen et~al.(2019)Justesen, Bontrager, Togelius, and
  Risi]{justesen2019deep}
Niels Justesen, Philip Bontrager, Julian Togelius, and Sebastian Risi.
\newblock Deep learning for video game playing.
\newblock \emph{IEEE Transactions on Games}, 12\penalty0 (1):\penalty0 1--20,
  2019.

\bibitem[Kaelbling et~al.(1996)Kaelbling, Littman, and
  Moore]{kaelbling1996reinforcement}
Leslie~Pack Kaelbling, Michael~L Littman, and Andrew~W Moore.
\newblock Reinforcement learning: A survey.
\newblock \emph{Journal of Artificial Intelligence Research}, 4:\penalty0
  237--285, 1996.

\bibitem[Kahneman(2011)]{kahneman2011thinking}
Daniel Kahneman.
\newblock \emph{Thinking, fast and slow}.
\newblock Farrar, Straus and Giroux, 2011.

\bibitem[Kaiser et~al.(2019)Kaiser, Babaeizadeh, Milos, Osinski, Campbell,
  Czechowski, Erhan, Finn, Kozakowski, Levine, et~al.]{kaiser2019model}
Lukasz Kaiser, Mohammad Babaeizadeh, Piotr Milos, Blazej Osinski, Roy~H
  Campbell, Konrad Czechowski, Dumitru Erhan, Chelsea Finn, Piotr Kozakowski,
  Sergey Levine, et~al.
\newblock Model-based reinforcement learning for {Atari}.
\newblock \emph{arXiv preprint arXiv:1903.00374}, 2019.

\bibitem[Kalweit and Boedecker(2017)]{kalweit2017uncertainty}
Gabriel Kalweit and Joschka Boedecker.
\newblock Uncertainty-driven imagination for continuous deep reinforcement
  learning.
\newblock In \emph{Conference on Robot Learning}, pages 195--206, 2017.

\bibitem[Kamyar and Taheri(2014)]{kamyar2014aircraft}
Reza Kamyar and Ehsan Taheri.
\newblock Aircraft optimal terrain/threat-based trajectory planning and
  control.
\newblock \emph{Journal of Guidance, Control, and Dynamics}, 37\penalty0
  (2):\penalty0 466--483, 2014.

\bibitem[Kansal and Martin(2018)]{learn}
Satwik Kansal and Brendan Martin.
\newblock Learn data science webpage., 2018.
\newblock URL
  \url{https://www.learndatasci.com/tutorials/reinforcement-q-learning-scratch-python-openai-gym/}.

\bibitem[Karl et~al.(2016)Karl, Soelch, Bayer, and Van~der Smagt]{karl2016deep}
Maximilian Karl, Maximilian Soelch, Justin Bayer, and Patrick Van~der Smagt.
\newblock Deep variational {Bayes} filters: Unsupervised learning of state
  space models from raw data.
\newblock \emph{arXiv preprint arXiv:1605.06432}, 2016.

\bibitem[Kelley(1960)]{kelley1960gradient}
Henry~J Kelley.
\newblock Gradient theory of optimal flight paths.
\newblock \emph{American Rocket Society Journal}, 30\penalty0 (10):\penalty0
  947--954, 1960.

\bibitem[Kempka et~al.(2016)Kempka, Wydmuch, Runc, Toczek, and
  Ja{\'s}kowski]{kempka2016vizdoom}
Micha{\l} Kempka, Marek Wydmuch, Grzegorz Runc, Jakub Toczek, and Wojciech
  Ja{\'s}kowski.
\newblock {VizDoom: A Doom-based} {AI} research platform for visual
  reinforcement learning.
\newblock In \emph{2016 IEEE Conference on Computational Intelligence and
  Games}, pages 1--8, 2016.

\bibitem[Kingma and Welling(2013)]{kingma2013auto}
Diederik~P Kingma and Max Welling.
\newblock Auto-encoding variational {Bayes}.
\newblock \emph{arXiv preprint arXiv:1312.6114}, 2013.

\bibitem[Kingma and Welling(2019)]{kingma2019introduction}
Diederik~P Kingma and Max Welling.
\newblock An introduction to variational autoencoders.
\newblock \emph{arXiv preprint arXiv:1906.02691}, 2019.

\bibitem[Kober et~al.(2013)Kober, Bagnell, and Peters]{kober2013reinforcement}
Jens Kober, J~Andrew Bagnell, and Jan Peters.
\newblock Reinforcement learning in robotics: A survey.
\newblock \emph{The International Journal of Robotics Research}, 32\penalty0
  (11):\penalty0 1238--1274, 2013.

\bibitem[Konda and Tsitsiklis(2000)]{konda2000actor}
Vijay~R Konda and John~N Tsitsiklis.
\newblock Actor--critic algorithms.
\newblock In \emph{Advances in Neural Information Processing Systems}, pages
  1008--1014, 2000.

\bibitem[Kwon et~al.(1983)Kwon, Bruckstein, and Kailath]{kwon1983stabilizing}
W~Hi Kwon, AM~Bruckstein, and T~Kailath.
\newblock Stabilizing state-feedback design via the moving horizon method.
\newblock \emph{International Journal of Control}, 37\penalty0 (3):\penalty0
  631--643, 1983.

\bibitem[Lakshminarayanan et~al.(2017)Lakshminarayanan, Pritzel, and
  Blundell]{lakshminarayanan2017simple}
Balaji Lakshminarayanan, Alexander Pritzel, and Charles Blundell.
\newblock Simple and scalable predictive uncertainty estimation using deep
  ensembles.
\newblock In \emph{Advances in Neural Information Processing Systems}, pages
  6402--6413, 2017.

\bibitem[LeCun et~al.(2015)LeCun, Bengio, and Hinton]{lecun2015deep}
Yann LeCun, Yoshua Bengio, and Geoffrey Hinton.
\newblock Deep learning.
\newblock \emph{Nature}, 521\penalty0 (7553):\penalty0 436, 2015.

\bibitem[Levine and Abbeel(2014)]{levine2014learning}
Sergey Levine and Pieter Abbeel.
\newblock Learning neural network policies with guided policy search under
  unknown dynamics.
\newblock In \emph{Advances in Neural Information Processing Systems}, pages
  1071--1079, 2014.

\bibitem[Levine and Koltun(2013)]{levine2013guided}
Sergey Levine and Vladlen Koltun.
\newblock Guided policy search.
\newblock In \emph{International Conference on Machine Learning}, pages 1--9,
  2013.

\bibitem[Mnih et~al.(2013)Mnih, Kavukcuoglu, Silver, Graves, Antonoglou,
  Wierstra, and Riedmiller]{mnih2013playing}
Volodymyr Mnih, Koray Kavukcuoglu, David Silver, Alex Graves, Ioannis
  Antonoglou, Daan Wierstra, and Martin Riedmiller.
\newblock Playing {A}tari with deep reinforcement learning.
\newblock \emph{arXiv preprint arXiv:1312.5602}, 2013.

\bibitem[Mnih et~al.(2015)Mnih, Kavukcuoglu, Silver, Rusu, Veness, Bellemare,
  Graves, Riedmiller, Fidjeland, Ostrovski, et~al.]{mnih2015human}
Volodymyr Mnih, Koray Kavukcuoglu, David Silver, Andrei~A Rusu, Joel Veness,
  Marc~G Bellemare, Alex Graves, Martin Riedmiller, Andreas~K Fidjeland, Georg
  Ostrovski, et~al.
\newblock Human-level control through deep reinforcement learning.
\newblock \emph{Nature}, 518\penalty0 (7540):\penalty0 529, 2015.

\bibitem[Mnih et~al.(2016)Mnih, Badia, Mirza, Graves, Lillicrap, Harley,
  Silver, and Kavukcuoglu]{mnih2016asynchronous}
Volodymyr Mnih, Adria~Puigdomenech Badia, Mehdi Mirza, Alex Graves, Timothy
  Lillicrap, Tim Harley, David Silver, and Koray Kavukcuoglu.
\newblock Asynchronous methods for deep reinforcement learning.
\newblock In \emph{International Conference on Machine Learning}, pages
  1928--1937, 2016.

\bibitem[Moerland et~al.(2018)Moerland, Broekens, Plaat, and
  Jonker]{moerland2018a0c}
Thomas~M Moerland, Joost Broekens, Aske Plaat, and Catholijn~M Jonker.
\newblock A0c: Alpha zero in continuous action space.
\newblock \emph{arXiv preprint arXiv:1805.09613}, 2018.

\bibitem[Moerland et~al.(2020{\natexlab{a}})Moerland, Broekens, and
  Jonker]{moerland2020framework}
Thomas~M Moerland, Joost Broekens, and Catholijn~M Jonker.
\newblock A framework for reinforcement learning and planning.
\newblock \emph{arXiv preprint arXiv:2006.15009}, 2020{\natexlab{a}}.

\bibitem[Moerland et~al.(2020{\natexlab{b}})Moerland, Broekens, and
  Jonker]{moerland2020model}
Thomas~M Moerland, Joost Broekens, and Catholijn~M Jonker.
\newblock Model-based reinforcement learning: A survey.
\newblock \emph{arXiv preprint arXiv:2006.16712}, 2020{\natexlab{b}}.

\bibitem[Nagabandi et~al.(2018)Nagabandi, Kahn, Fearing, and
  Levine]{nagabandi2018neural}
Anusha Nagabandi, Gregory Kahn, Ronald~S Fearing, and Sergey Levine.
\newblock Neural network dynamics for model-based deep reinforcement learning
  with model-free fine-tuning.
\newblock In \emph{2018 IEEE International Conference on Robotics and
  Automation (ICRA)}, pages 7559--7566, 2018.

\bibitem[Nardelli et~al.(2018)Nardelli, Synnaeve, Lin, Kohli, Torr, and
  Usunier]{nardelli2018value}
Nantas Nardelli, Gabriel Synnaeve, Zeming Lin, Pushmeet Kohli, Philip~HS Torr,
  and Nicolas Usunier.
\newblock Value propagation networks.
\newblock \emph{arXiv preprint arXiv:1805.11199}, 2018.

\bibitem[Oh et~al.(2015)Oh, Guo, Lee, Lewis, and Singh]{oh2015action}
Junhyuk Oh, Xiaoxiao Guo, Honglak Lee, Richard~L Lewis, and Satinder Singh.
\newblock Action-conditional video prediction using deep networks in {A}tari
  games.
\newblock In \emph{Advances in Neural Information Processing Systems}, pages
  2863--2871, 2015.

\bibitem[Oh et~al.(2017)Oh, Singh, and Lee]{oh2017value}
Junhyuk Oh, Satinder Singh, and Honglak Lee.
\newblock Value prediction network.
\newblock In \emph{Advances in Neural Information Processing Systems}, pages
  6118--6128, 2017.

\bibitem[Ontan{\'o}n et~al.(2013)Ontan{\'o}n, Synnaeve, Uriarte, Richoux,
  Churchill, and Preuss]{ontanon2013survey}
Santiago Ontan{\'o}n, Gabriel Synnaeve, Alberto Uriarte, Florian Richoux, David
  Churchill, and Mike Preuss.
\newblock A survey of real-time strategy game {AI} research and competition in
  {StarCraft}.
\newblock \emph{IEEE Transactions on Computational Intelligence and AI in
  Games}, 5\penalty0 (4):\penalty0 293--311, 2013.

\bibitem[Pascanu et~al.(2017)Pascanu, Li, Vinyals, Heess, Buesing,
  Racani{\`e}re, Reichert, Weber, Wierstra, and Battaglia]{pascanu2017learning}
Razvan Pascanu, Yujia Li, Oriol Vinyals, Nicolas Heess, Lars Buesing, Sebastien
  Racani{\`e}re, David Reichert, Th{\'e}ophane Weber, Daan Wierstra, and Peter
  Battaglia.
\newblock Learning model-based planning from scratch.
\newblock \emph{arXiv preprint arXiv:1707.06170}, 2017.

\bibitem[Plaat(2020)]{plaat2020learning}
Aske Plaat.
\newblock \emph{Learning to Play: Reinforcement Learning and Games}.
\newblock Springer Verlag, Heidelberg, See \url{https://learningtoplay.net},
  2020.

\bibitem[Polydoros and Nalpantidis(2017)]{polydoros2017survey}
Athanasios~S Polydoros and Lazaros Nalpantidis.
\newblock Survey of model-based reinforcement learning: Applications on
  robotics.
\newblock \emph{Journal of Intelligent \& Robotic Systems}, 86\penalty0
  (2):\penalty0 153--173, 2017.

\bibitem[Richards(2005)]{richards2005robust}
Arthur~George Richards.
\newblock \emph{Robust constrained model predictive control}.
\newblock PhD thesis, Massachusetts Institute of Technology, 2005.

\bibitem[Risi and Preuss(2020)]{risi2020chess}
Sebastian Risi and Mike Preuss.
\newblock {From Chess and Atari to StarCraft and Beyond: How Game AI is Driving
  the World of AI}.
\newblock \emph{KI-K{\"u}nstliche Intelligenz}, pages 1--11, 2020.

\bibitem[Rosin(2011)]{rosin2011multi}
Christopher~D Rosin.
\newblock Multi-armed bandits with episode context.
\newblock \emph{Annals of Mathematics and Artificial Intelligence}, 61\penalty0
  (3):\penalty0 203--230, 2011.

\bibitem[Schleich et~al.(2019)Schleich, Klamt, and Behnke]{schleich2019value}
Daniel Schleich, Tobias Klamt, and Sven Behnke.
\newblock Value iteration networks on multiple levels of abstraction.
\newblock \emph{arXiv preprint arXiv:1905.11068}, 2019.

\bibitem[Schmidhuber(1990{\natexlab{a}})]{schmidhuber1990line}
J{\"u}rgen Schmidhuber.
\newblock An on-line algorithm for dynamic reinforcement learning and planning
  in reactive environments.
\newblock In \emph{1990 IJCNN International Joint Conference on Neural
  Networks}, pages 253--258. IEEE, 1990{\natexlab{a}}.

\bibitem[Schmidhuber(1990{\natexlab{b}})]{schmidhuber1990making}
J\"urgen Schmidhuber.
\newblock Making the world differentiable: On using self-supervised fully
  recurrent neural networks for dynamic reinforcement learning and planning in
  non-stationary environments.
\newblock 1990{\natexlab{b}}.

\bibitem[Schrittwieser et~al.(2020)Schrittwieser, Antonoglou, Hubert, Simonyan,
  Sifre, Schmitt, Guez, Lockhart, Hassabis, Graepel,
  et~al.]{schrittwieser2020mastering}
Julian Schrittwieser, Ioannis Antonoglou, Thomas Hubert, Karen Simonyan,
  Laurent Sifre, Simon Schmitt, Arthur Guez, Edward Lockhart, Demis Hassabis,
  Thore Graepel, et~al.
\newblock Mastering atari, go, chess and shogi by planning with a learned
  model.
\newblock \emph{Nature}, 588\penalty0 (7839):\penalty0 604--609, 2020.

\bibitem[Schulman et~al.(2017)Schulman, Wolski, Dhariwal, Radford, and
  Klimov]{schulman2017proximal}
John Schulman, Filip Wolski, Prafulla Dhariwal, Alec Radford, and Oleg Klimov.
\newblock Proximal policy optimization algorithms.
\newblock \emph{arXiv preprint arXiv:1707.06347}, 2017.

\bibitem[Sekar et~al.(2020)Sekar, Rybkin, Daniilidis, Abbeel, Hafner, and
  Pathak]{sekar2020planning}
Ramanan Sekar, Oleh Rybkin, Kostas Daniilidis, Pieter Abbeel, Danijar Hafner,
  and Deepak Pathak.
\newblock Planning to explore via self-supervised world models.
\newblock \emph{arXiv preprint arXiv:2005.05960}, 2020.

\bibitem[Silver et~al.(2012)Silver, Sutton, and M{\"u}ller]{silver2012temporal}
David Silver, Richard~S Sutton, and Martin M{\"u}ller.
\newblock Temporal-difference search in computer {Go}.
\newblock \emph{Machine Learning}, 87\penalty0 (2):\penalty0 183--219, 2012.

\bibitem[Silver et~al.(2014)Silver, Lever, Heess, Degris, Wierstra, and
  Riedmiller]{silver2014deterministic}
David Silver, Guy Lever, Nicolas Heess, Thomas Degris, Daan Wierstra, and
  Martin Riedmiller.
\newblock Deterministic policy gradient algorithms.
\newblock 2014.

\bibitem[Silver et~al.(2016)Silver, Huang, Maddison, Guez, Sifre, van~den
  Driessche, Schrittwieser, Antonoglou, Panneershelvam, Lanctot, Dieleman,
  Grewe, Nham, Kalchbrenner, Sutskever, Lillicrap, Leach, Kavukcuoglu, Graepel,
  and Hassabis]{silver2016mastering}
David Silver, Aja Huang, Chris~J. Maddison, Arthur Guez, Laurent Sifre, George
  van~den Driessche, Julian Schrittwieser, Ioannis Antonoglou, Veda
  Panneershelvam, Marc Lanctot, Sander Dieleman, Dominik Grewe, John Nham, Nal
  Kalchbrenner, Ilya Sutskever, Timothy Lillicrap, Madeleine Leach, Koray
  Kavukcuoglu, Thore Graepel, and Demis Hassabis.
\newblock Mastering the game of {Go} with deep neural networks and tree search.
\newblock \emph{Nature}, 529\penalty0 (7587):\penalty0 484, 2016.

\bibitem[Silver et~al.(2017{\natexlab{a}})Silver, Schrittwieser, Simonyan,
  Antonoglou, Huang, Guez, Hubert, Baker, Lai, Bolton, Chen, Lillicrap, Hui,
  Sifre, van~den Driessche, Graepel, and Hassabis]{silver2017mastering}
David Silver, Julian Schrittwieser, Karen Simonyan, Ioannis Antonoglou, Aja
  Huang, Arthur Guez, Thomas Hubert, Lucas Baker, Matthew Lai, Adrian Bolton,
  Yutian Chen, Timothy Lillicrap, Fan Hui, Laurent Sifre, George van~den
  Driessche, Thore Graepel, and Demis Hassabis.
\newblock Mastering the game of {Go} without human knowledge.
\newblock \emph{Nature}, 550\penalty0 (7676):\penalty0 354, 2017{\natexlab{a}}.

\bibitem[Silver et~al.(2017{\natexlab{b}})Silver, van Hasselt, Hessel, Schaul,
  Guez, Harley, Dulac-Arnold, Reichert, Rabinowitz, Barreto,
  et~al.]{silver2017predictron}
David Silver, Hado van Hasselt, Matteo Hessel, Tom Schaul, Arthur Guez, Tim
  Harley, Gabriel Dulac-Arnold, David Reichert, Neil Rabinowitz, Andre Barreto,
  et~al.
\newblock The predictron: End-to-end learning and planning.
\newblock In \emph{Proceedings of the 34th International Conference on Machine
  Learning}, pages 3191--3199, 2017{\natexlab{b}}.

\bibitem[Silver et~al.(2018)Silver, Hubert, Schrittwieser, Antonoglou, Lai,
  Guez, Lanctot, Sifre, Kumaran, Graepel, Lillicrap, Simonyan, and
  Hassabis]{silver2018general}
David Silver, Thomas Hubert, Julian Schrittwieser, Ioannis Antonoglou, Matthew
  Lai, Arthur Guez, Marc Lanctot, Laurent Sifre, Dharshan Kumaran, Thore
  Graepel, Timothy Lillicrap, Karen Simonyan, and Demis Hassabis.
\newblock A general reinforcement learning algorithm that masters chess, shogi,
  and {Go} through self-play.
\newblock \emph{Science}, 362\penalty0 (6419):\penalty0 1140--1144, 2018.

\bibitem[Srinivas et~al.(2018)Srinivas, Jabri, Abbeel, Levine, and
  Finn]{srinivas2018universal}
Aravind Srinivas, Allan Jabri, Pieter Abbeel, Sergey Levine, and Chelsea Finn.
\newblock Universal planning networks.
\newblock \emph{arXiv preprint arXiv:1804.00645}, 2018.

\bibitem[Sutton(1990)]{sutton1990integrated}
Richard~S Sutton.
\newblock Integrated architectures for learning, planning, and reacting based
  on approximating dynamic programming.
\newblock In \emph{Machine learning proceedings 1990}, pages 216--224.
  Elsevier, 1990.

\bibitem[Sutton(1991)]{sutton1991dyna}
Richard~S Sutton.
\newblock Dyna, an integrated architecture for learning, planning, and
  reacting.
\newblock \emph{ACM Sigart Bulletin}, 2\penalty0 (4):\penalty0 160--163, 1991.

\bibitem[Sutton and Barto(2018)]{sutton2018introduction}
Richard~S Sutton and Andrew~G Barto.
\newblock \emph{Reinforcement learning, An Introduction, Second Edition}.
\newblock MIT Press, 2018.

\bibitem[Talvitie(2015)]{talvitie2015agnostic}
Erik Talvitie.
\newblock Agnostic system identification for monte carlo planning.
\newblock In \emph{Twenty-Ninth AAAI Conference on Artificial Intelligence},
  2015.

\bibitem[Tamar et~al.(2016)Tamar, Wu, Thomas, Levine, and
  Abbeel]{tamar2016value}
Aviv Tamar, Yi~Wu, Garrett Thomas, Sergey Levine, and Pieter Abbeel.
\newblock Value iteration networks.
\newblock In \emph{Adv. in Neural Information Processing Systems}, pages
  2154--2162, 2016.

\bibitem[Tassa et~al.(2012)Tassa, Erez, and Todorov]{tassa2012synthesis}
Yuval Tassa, Tom Erez, and Emanuel Todorov.
\newblock Synthesis and stabilization of complex behaviors through online
  trajectory optimization.
\newblock In \emph{2012 IEEE/RSJ International Conference on Intelligent Robots
  and Systems}, pages 4906--4913, 2012.

\bibitem[Tassa et~al.(2018)Tassa, Doron, Muldal, Erez, Li, Casas, Budden,
  Abdolmaleki, Merel, Lefrancq, et~al.]{tassa2018deepmind}
Yuval Tassa, Yotam Doron, Alistair Muldal, Tom Erez, Yazhe Li, Diego de~Las
  Casas, David Budden, Abbas Abdolmaleki, Josh Merel, Andrew Lefrancq, et~al.
\newblock Deepmind control suite.
\newblock \emph{arXiv preprint arXiv:1801.00690}, 2018.

\bibitem[Todorov et~al.(2012)Todorov, Erez, and Tassa]{todorov2012mujoco}
Emanuel Todorov, Tom Erez, and Yuval Tassa.
\newblock {MuJoCo}: A physics engine for model-based control.
\newblock In \emph{IEEE/RSJ International Conference on Intelligent Robots and
  Systems (IROS)}, pages 5026--5033, 2012.

\bibitem[Torrado et~al.(2018)Torrado, Bontrager, Togelius, Liu, and
  Perez-Liebana]{torrado2018deep}
Ruben~Rodriguez Torrado, Philip Bontrager, Julian Togelius, Jialin Liu, and
  Diego Perez-Liebana.
\newblock Deep reinforcement learning for general video game ai.
\newblock In \emph{2018 IEEE Conference on Computational Intelligence and Games
  (CIG)}, pages 1--8. IEEE, 2018.

\bibitem[Van Der~Maaten et~al.(2009)Van Der~Maaten, Postma, Van~den Herik,
  et~al.]{van2009dimensionality}
Laurens Van Der~Maaten, Eric Postma, Jaap Van~den Herik, et~al.
\newblock Dimensionality reduction: a comparative.
\newblock \emph{J Mach Learn Res}, 10\penalty0 (66-71):\penalty0 13, 2009.

\bibitem[Vinyals et~al.(2017)Vinyals, Ewalds, Bartunov, Georgiev, Vezhnevets,
  Yeo, Makhzani, K{\"u}ttler, Agapiou, Schrittwieser,
  et~al.]{vinyals2017starcraft}
Oriol Vinyals, Timo Ewalds, Sergey Bartunov, Petko Georgiev, Alexander~Sasha
  Vezhnevets, Michelle Yeo, Alireza Makhzani, Heinrich K{\"u}ttler, John
  Agapiou, Julian Schrittwieser, et~al.
\newblock {Starcraft II}: A new challenge for reinforcement learning.
\newblock \emph{arXiv preprint arXiv:1708.04782}, 2017.

\bibitem[Vinyals et~al.(2019)Vinyals, Babuschkin, Czarnecki, Mathieu, Dudzik,
  Chung, Choi, Powell, Ewalds, Georgiev, et~al.]{vinyals2019grandmaster}
Oriol Vinyals, Igor Babuschkin, Wojciech~M Czarnecki, Micha{\"e}l Mathieu,
  Andrew Dudzik, Junyoung Chung, David~H Choi, Richard Powell, Timo Ewalds,
  Petko Georgiev, et~al.
\newblock Grandmaster level in starcraft ii using multi-agent reinforcement
  learning.
\newblock \emph{Nature}, 575\penalty0 (7782):\penalty0 350--354, 2019.

\bibitem[Wang et~al.(2019)Wang, Bao, Clavera, Hoang, Wen, Langlois, Zhang,
  Zhang, Abbeel, and Ba]{wang2019benchmarking}
Tingwu Wang, Xuchan Bao, Ignasi Clavera, Jerrick Hoang, Yeming Wen, Eric
  Langlois, Shunshi Zhang, Guodong Zhang, Pieter Abbeel, and Jimmy Ba.
\newblock Benchmarking model-based reinforcement learning.
\newblock \emph{preprint arXiv:1907.02057}, 2019.

\bibitem[Watkins(1989)]{watkins1989learning}
Christopher~JCH Watkins.
\newblock \emph{Learning from delayed rewards}.
\newblock PhD thesis, King's College, Cambridge, 1989.

\bibitem[Weber et~al.(2017)Weber, Racani{\`e}re, Reichert, Buesing, Guez,
  Rezende, Badia, Vinyals, Heess, Li, et~al.]{racaniere2017imagination}
Th{\'e}ophane Weber, S{\'e}bastien Racani{\`e}re, David Reichert, Lars Buesing,
  Arthur Guez, Danilo~Jimenez Rezende, Adria~Puigdomenech Badia, Oriol Vinyals,
  Nicolas Heess, Yujia Li, et~al.
\newblock Imagination-augmented agents for deep reinforcement learning.
\newblock In \emph{Advances in Neural Information Processing Systems}, pages
  5690--5701, 2017.

\bibitem[Wong et~al.(2021)Wong, B\"ack, Kononova, and Plaat]{wong2021survey}
Annie Wong, Thomas B\"ack, Anna~V. Kononova, and Aske Plaat.
\newblock Multiagent deep reinforcement learning: Challenges and directions
  towards human-like approaches.
\newblock \emph{Artificial Intelligence Review}, 2021.

\bibitem[Xingjian et~al.(2015)Xingjian, Chen, Wang, Yeung, Wong, and
  Woo]{xingjian2015convolutional}
SHI Xingjian, Zhourong Chen, Hao Wang, Dit-Yan Yeung, Wai-Kin Wong, and
  Wang-chun Woo.
\newblock Convolutional {LSTM} network: A machine learning approach for
  precipitation nowcasting.
\newblock In \emph{Advances in Neural Information Processing Systems}, pages
  802--810, 2015.

\bibitem[Zambaldi et~al.(2018)Zambaldi, Raposo, Santoro, Bapst, Li, Babuschkin,
  Tuyls, Reichert, Lillicrap, Lockhart, et~al.]{zambaldi2018relational}
Vinicius Zambaldi, David Raposo, Adam Santoro, Victor Bapst, Yujia Li, Igor
  Babuschkin, Karl Tuyls, David Reichert, Timothy Lillicrap, Edward Lockhart,
  et~al.
\newblock Relational deep reinforcement learning.
\newblock \emph{arXiv preprint arXiv:1806.01830}, 2018.

\end{thebibliography}

\end{document}